\documentclass[journal]{IEEEtran}

\usepackage{cite}
\usepackage{amsmath,amssymb,amsfonts}
\usepackage{algorithmic}
\usepackage{graphicx}
\usepackage{textcomp}
\usepackage{xcolor}
\usepackage{times}
\usepackage{soul}
\usepackage[utf8]{inputenc}
\usepackage{amsmath}
\usepackage{amsfonts}
\usepackage{multirow}
\usepackage{amsthm}

\usepackage{algorithm} 
\usepackage{caption,subcaption}
\usepackage[justification=centering]{subcaption}

\usepackage{color}
\usepackage{mathtools}
\usepackage{mathptmx}
\usepackage{float}
\usepackage{color}
\usepackage{makecell}
\usepackage{epstopdf}
\usepackage{colortbl}
\usepackage{pdflscape}
\usepackage{booktabs}
\usepackage{threeparttable}
\usepackage{makecell}

\def\BibTeX{{\rm B\kern-.05em{\sc i\kern-.025em b}\kern-.08em
		T\kern-.1667em\lower.7ex\hbox{E}\kern-.125emX}} 
\begin{document}

\title{Face Hallucination with Finishing Touches}

\author{Yang Zhang,
        Ivor W. Tsang,
        Jun Li,
        Ping Liu,
        Xiaobo Lu,
        and~Xin~Yu
\thanks{This work was done when the first author visited the University of Technology Sydney. Y. Zhang is with the School of Automation, Southeast University, Nanjing 210096, China; the Australian Artificial Intelligence Institute, University of Technology Sydney, Ultimo, NSW 2007, Australia; the Key Laboratory of Measurement and Control of Complex Systems of Engineering, Ministry of Education, Nanjing 210096, China, e-mail: zhangyang201703@126.com.

I. W. Tsang, J. Li and Xin Yu are with the Australian Artificial Intelligence Institute, University of Technology Sydney, Ultimo, NSW 2007, Australia, e-mail: (ivor.tsang@uts.edu.au, Jun.Li@uts.edu.au, and xin.yu@uts.edu.au).

P. Liu is with the Institute of High Performance Computing, Research Agency for Science, Technology and Research (A*STAR), Singapore 138634 (e-mail: pino.pingliu@gmail.com).

X. B. Lu is with the School of Automation, Southeast University, Nanjing
210096, China; Key Laboratory of Measurement and Control of Complex
Systems of Engineering, Ministry of Education, Nanjing 210096, China, email: xblu2013@126.com.

This work was done when Xin Yu was with the Research School of Engineering, Australian National University, Canberra, Australia.
}
}

\markboth{Journal of \LaTeX\ Class Files,~Vol.~14, No.~8, August~2015}%
{Shell \MakeLowercase{\textit{et al.}}: Bare Demo of IEEEtran.cls for IEEE Journals}

\maketitle

\begin{abstract}
Obtaining a high-quality frontal face image from a low-resolution (LR) non-frontal face image is primarily important for many facial analysis applications.
However, mainstreams either focus on super-resolving near-frontal LR faces or frontalizing non-frontal high-resolution (HR) faces.
It is desirable to perform both tasks seamlessly for daily-life unconstrained face images.
In this paper, we present a novel Vivid Face Hallucination Generative Adversarial Network (VividGAN) for simultaneously super-resolving and frontalizing tiny non-frontal face images.
VividGAN consists of coarse-level and fine-level Face Hallucination Networks (FHnet) and two discriminators, \emph{i.e.,} Coarse-D and Fine-D.
The coarse-level FHnet generates a frontal coarse HR face and then the fine-level FHnet makes use of the facial component appearance prior, \emph{i.e.,} fine-grained facial components, to attain a frontal HR face image with authentic details.
In the fine-level FHnet, we also design a facial component-aware module that adopts the facial geometry guidance as clues to accurately align and merge the frontal coarse HR face and prior information.
Meanwhile, two-level discriminators are designed to capture both the global outline of a face image as well as detailed facial characteristics.
The Coarse-D enforces the coarsely hallucinated faces to be upright and complete while the Fine-D focuses on the fine hallucinated ones for sharper details.
Extensive experiments demonstrate that our VividGAN achieves photo-realistic frontal HR faces, reaching superior performance in downstream tasks, \emph{i.e.,} face recognition and expression classification, compared with other state-of-the-art methods.
\end{abstract}


\IEEEpeerreviewmaketitle

\section{Introduction}
As the requirement of public security increases, authentication plays an important role in daily life~\cite{zhao2003face}.
High-quality face images are one of the most discriminative biometric cues to provide identity verification.
{\color{black}{However, in most imaging conditions, long distance and relative location between public cameras and human faces are inevitable interference factors, leading to a vast collection of non-frontal tiny face images.
Such low-quality face images not only impede human observation but also degrade the performance of downstream machine perception algorithms~\cite{liu2020point, liu2020omni}}}

\begin{figure}[t]
\centering
\vspace{-0.1cm}
\includegraphics[height=10cm,width=0.49\textwidth]{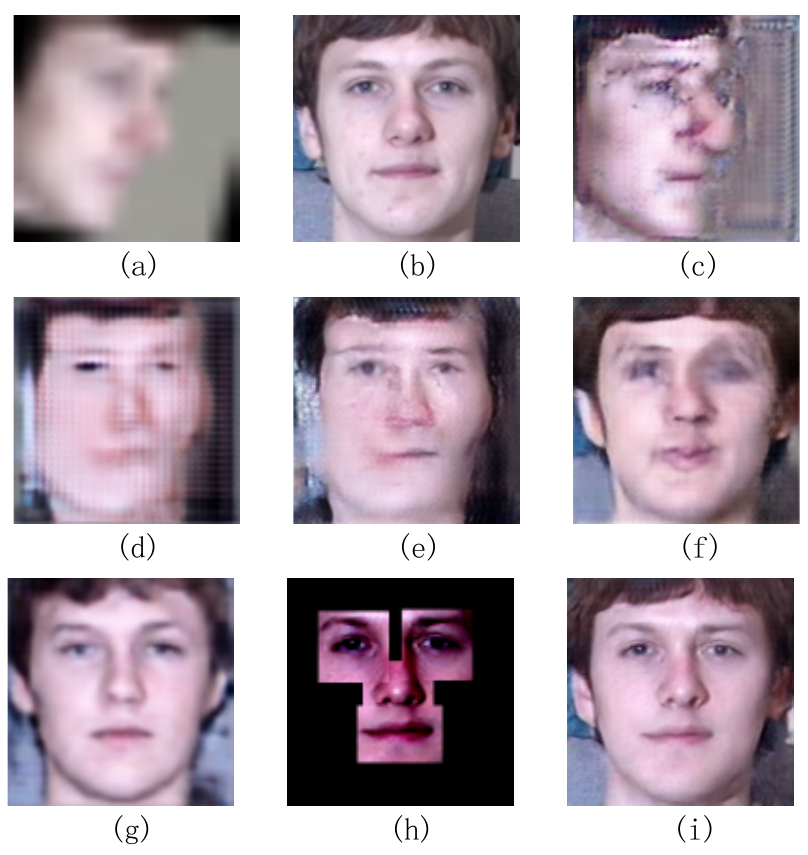}
\caption{Face super-resolution and frontalization results on a testing non-frontal LR face.
(a) An input non-frontal LR face ($16\times 16$ pixels). (b) The ground-truth frontal face ($128\times 128$ pixels, not available in training). (c) Result obtained by applying~\cite{yu2018face} to (a). (d) Face frontalization result of (a) by applying~\cite{tran2018representation} after bicubic upsampling and alignment, where a single STN network~\cite{jaderberg2015spatial} is introduced to align (a) first. (e) Result obtained by applying~\cite{tran2018representation} first and then~\cite{yu2018face}. (f) Result obtained by applying~\cite{yu2018face} first and then~\cite{tran2018representation}. (g) Result of TANN~\cite{yu2019can}. (h) Fine-grained facial components. (i) Result of our VividGAN.}
\label{fig0}
\vspace{-0.6cm}
\end{figure}

Motivated by this challenge, many researchers resort to face super-resolution (SR) techniques to recover high-resolution (HR) face images from the tiny inputs.
{\color{black}{State-of-the-art face SR methods~\cite{zhu2016deep,cao2017attention,chen2018fsrnet,yu2018imagining,yu2018face} utilize deep convolutional neural networks to super-resolve near-frontal low-resolution (LR) faces.
Meanwhile, some works~\cite{thies2016face2face,sagonas2015robust,hassner2015effective,huang2017beyond} have been proposed to frontalize faces from their non-frontal counterparts.
However, those face frontalization techniques~\cite{blanz1999morphable,zhu2015high} mainly warp 2D HR face images based on real-time detected facial landmarks. Hence, they might fail to frontalize very tiny non-frontal faces.}}

{\color{black}{In this work, we aim to hallucinate a frontal HR face from a non-frontal LR input.
A naive idea is to \textit{sequentially} combine existing face SR and frontalization models.}}
However, as seen in Figs.~\ref{fig0}(e) and (f), the results undergo obvious distortions and severe artifacts.
This is mainly caused by two factors:
$\left(\textbf{\romannumeral1}\right)$ Both face SR and face frontalizaiton are ill-posed inverse problems, and the errors in either process cannot be eliminated by the other one but are exaggerated.
{\color{black}{$\left(\textbf{\romannumeral2}\right)$ Pose variations bring challenges to state-of-the-art face SR techniques (see Fig.~\ref{fig0}(c)) and low resolution images impose difficulties on existing face frontalization algorithms (see Fig.~\ref{fig0}(d)). As a result, we face a chicken-and-egg problem: face SR would be facilitated by face frontalization, while the latter requires an HR face. This challenging issue cannot be addressed by simply combining those two methods.}}

Very recently, the work~\cite{yu2019can} proposes a Transformative Adversarial Neural Network (TANN) to jointly frontalize and super-resolve non-frontal LR face images.
{\color{black}Unlike the direct combination of face frontalization and SR models, the joint mechanism performed in TANN is able to avoid artifacts.
However, since TANN is a single-stage face hallucination network, it does not process a “looking back” mechanism to modify upsampled faces. 
Therefore, TANN may output blurry facial details under extreme poses or challenging expressions.
As visible in Fig.~\ref{fig0}(g), unnatural artifacts still exist in the result of TANN.}

\begin{figure}[t]
\centering
\vspace{-0.1cm}
\includegraphics[height=2cm,width=0.49\textwidth]{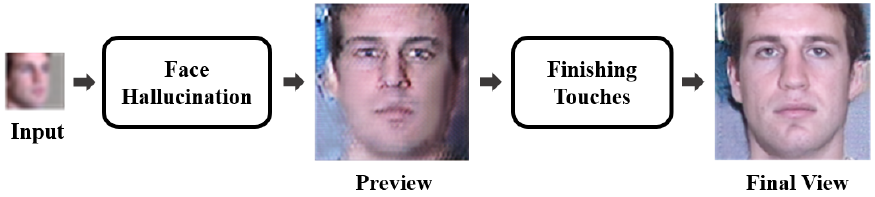}
\caption{The pipeline of our VividGAN.}
\label{fig2}
\vspace{-0.6cm}
\end{figure}

Different from previous works, we propose a novel Vivid Face Hallucination Generative Adversarial Network (VividGAN).
{\color{black}Our VividGAN \textit{jointly} super-resolves and frontalizes tiny non-frontal face images and is designed to \textit{progressively} hallucinate a non-frontal LR face in a coarse-to-fine manner (see Fig.~\ref{fig2}), thus reducing unwanted blurriness and artifacts significantly.
In this manner, these two tasks, \emph{i.e.,} face SR and frontalization, facilitate each other in a unified framework. 
Furthermore, we observe that facial components, such as eyes, noses and mouths, are the most distinguishable parts to humans.}
Therefore, we introduce the facial component appearance prior, \emph{i.e.,} fine-grained facial components, as the semantic guidance to achieve realistic facial details as our ``finishing touches".

Specifically, our VividGAN consists of a Vivid Face Hallucination Network (Vivid-FHnet) that comprises a coarse-level FH network and a fine-level FH network, as well as two discriminators, \emph{i.e.,} Coarse-D and Fine-D. Their details are described as follows:

$\left(\textbf{\romannumeral1}\right)$ A coarse-level FH network is designed to super-resolve and frontalize an input non-frontal LR face roughly.
We do not assume the input face is aligned in advance.
Instead, we interweave the spatial transformation networks (STNs)~\cite{jaderberg2015spatial} with upsampling layers to compensate for misalignments.
In order to guarantee the content-integrity of hallucinated results, we introduce a mirror symmetry loss based on domain-specific knowledge of human faces.
As a result, we can generate coarsely frontal HR faces, which facilitates the next hallucination procedure.

$\left(\textbf{\romannumeral2}\right)$ Subsequently, we propose a {\color{black}{fine-level FH network}} for {\color{black}{fine detail recovery}} of frontal HR face image by explicitly incorporating the {\color{black}{facial component appearance prior}}, \emph{i.e.,} fine-grained facial components.
{\color{black}{First, we design a touching-up subnetwork to effectively estimate the fine-grained facial components by integrating multi-scale upscaling and downsampling blocks (see Fig.~\ref{fig0}(h)).}}
Then, we feed the facial component appearance prior and the frontal coarse HR face into a fine-integration subnetwork for further refinement.
{\color{black}{Particularly, we design a facial component-aware module to align and merge the coarsely restored face and the facial component prior information.}}

$\left(\textbf{\romannumeral3}\right)$ Different from previous works~\cite{yu2019hallucinating,yu2018super,yu2019can,yu2020hallucinating,yu2019semantic} that use a single discriminator, {\color{black}{we employ two-level discriminators}}, which promote our coarse-to-fine Vivid-FHnet to produce photo-realistic results.
In this fashion, we are able to capture both global outlines and local details of faces.
Fig.~\ref{fig0}(i) illustrates that our hallucinated frontal HR face is more visually appealing than the results of the state-of-the-art methods.

In summary, our contributions are threefold: 
\begin{itemize}
\item We propose a novel framework, dubbed Vivid Face Hallucination Generative Adversarial Network (VividGAN), to jointly tackle face SR and face frontalization in a unified framework. 
In particular, we {\color{black}{first upsample and frontalize low-resolution faces in a coarse granularity}} and  {\color{black}{then take advantage of facial component appearance priors to remedy the details of coarsely upsampled frontal HR faces.}}
\item We propose a facial component-aware module for face enhancement and alignment. In this way, fine-grained facial components are incorporated into the frontal coarse HR face seamlessly.
\item Our experiments demonstrate that VividGAN is able to frontalize and super-resolve (by an upscaling factor of $8\times$) very {\color{black}{low-resolution}} face images (\emph{e.g.,} $16\times 16$) undergoing {\color{black}{large poses}} (\emph{e.g.,} $90^{o}$) and {\color{black}{complex expressions}} (\emph{e.g.,} ``disgust”, ``fear”). 
Moreover, {\color{black}{our VividGAN is also able to provide superior hallucinated face images for downstream tasks, \emph{i.e.,} face recognition and expression classification, in comparison to the state-of-the-art.}}
\end{itemize}

\section{Related Work} 
\subsection{Face Super-resolution}
Face Super-resolution (SR) aims at establishing the intensity relationships between input LR and output HR face images. 
The previous works are generally categorized into three mainstreams: {\color{black}{holistic-based, part-based, and deep learning based methods.}}

The basic principle of holistic-based techniques is to upsample a whole LR face by a global face model.
Wang~\textit{et al.}~\cite{wang2005hallucinating} formulate a linear mapping between LR and HR images to achieve face SR based on an Eigen-transformation of LR faces.
Liu~\textit{et al.}~\cite{liu2007face} incorporate a bilateral filtering to mitigate the ghosting artifacts.
Kolouri and Rohde~\cite{kolouri2015transport} morph HR faces from aligned LR ones based on optimal transport and subspace learning. 
However, they require LR inputs to be precisely aligned and reference HR faces to exhibit similar canonical poses and natural expressions.

To address pose and expression variations, part-based methods are proposed to make use of exemplar facial patches to upsample local facial regions instead of imposing global constraints.
The approaches~\cite{ma2010hallucinating,farrugia2017face,jiang2014face} super-resolve local LR patches based on a weighted sum of exemplar facial patches in reference HR database.
Liu~\textit{et al.}~\cite{liu2017robust} develops a locality-constrained bi-layer network to jointly super-resolve LR faces as well as eliminate noise and outliers.
Moreover, SIFT flow~\cite{tappen2012bayesian} and facial landmarks~\cite{yang2018hallucinating} are introduced to locate facial components for further super-resolution.
Since these techniques need to localize facial components in LR inputs preciously, they may fail to process very LR faces.

Recently, deep learning based face SR methods have been actively explored and achieved superior performance compared to traditional methods.
Yu \textit{et al.}~\cite{yu2016ultra,yu2017face} develop GAN-based models to hallucinate very LR face images.
Huang \textit{et al.}~\cite{huang2019wavelet} incorporate the wavelet coefficients into deep convolutional networks to super-resolve LR inputs with multiple upscaling factors. 
Cao \textit{et al.}~\cite{cao2017attention} design an attention-aware mechanism and a local enhancement network to alternately enhance facial regions in super-resolution.
Xu \textit{et al.}~\cite{xu2017learning} jointly super-resolve and deblur face and text images with a multi-class adversarial loss. 
Dahl \textit{et al.}~\cite{dahl2017pixel} present an autoregressive Pixel-RNN~\cite{oord2016pixel} to hallucinate pre-aligned LR faces.
Yu \textit{et al.}~\cite{yu2019hallucinating} present a multiscale transformative discriminative network to hallucinate unaligned input LR face images with different resolutions.
Zhang \textit{et al.}~\cite{zhang2020copy} develop a two-branch super-resolution network to compensate and upsample ill-illuminated LR face images.
However, these methods focus on super-resolving near-frontal LR faces. 
Thus, they are restricted to the inputs under small pose variations.

Several face SR techniques have been proposed to super-resolve LR faces under large pose variations by introducing facial prior information~\cite{chen2018fsrnet,bulat2018learn,yu2018face}. 
Chen \textit{et al.}~\cite{chen2018fsrnet} incorporate facial geometry priors into their SR model to super-resolve LR faces.
Yu \textit{et al.}~\cite{yu2018face} exploit the facial component information from the intermediate upsampled features to encourage the upsampling stream to produce photo-realistic HR faces.
However, these techniques only super-resolve non-frontal LR faces {\color{black}{without frontalizing them for better visual perception.}}

\subsection{Face Frontalization}
Frontal view synthesis, termed as face frontalization, is a challenging task for its ill-posed nature, such as self-occlusions and pose variations.
Conventional and emerging researches on face frontalization can be grouped into three classes: {\color{black}{2D/3D local feature warping, statistic modeling as well as deep learning based methods.}}

The first category researches date back to the 3D Morphable Model (3DMM)~\cite{blanz1999morphable}, which extracts the shape and texture bases of a face in PCA subspace. 
Driven by 3DMM, Yang~\textit{et al.}~\cite{yang2011expression} formulate new expressions of input faces by estimating 3D surface from face appearance.
Meanwhile, approaches~\cite{jiang2005efficient,asthana2011fully,asthana2011fully,hassner2013viewing,taigman2014deepface,zhu2015high,masi2016we} generate frontal faces by mapping a non-frontal face onto a 3D reference surface mesh. 
However, detecting facial landmarks~\cite{blanz2003face} is the fundamental prerequisite for these approaches to determine the transformation between the query image and template.
Thus, they fail to address the face images with extreme yaw angles or in low resolutions.

Considering that frontal faces has the minimum rank of all various poses, another face frontalization research stream focuses on infering frontal views by solving a low-rank constrained minimization problem of statistical models. 
Sagonas \textit{et al.}~\cite{sagonas2015robust} achieve joint face frontalization and facial landmark detection in a whole framework.
However, their results cannot ensure that the frontalized faces are consistent with the ground-truth frontal ones.

More recently, deep learning based methods~\cite{zhu2014recover,yim2015rotating,cole2017face,hu2018pose,zhao2019multi,tran2018representation,thies2016face2face,huang2017beyond} have been leveraged for face frontalization research. 
Kan \textit{et al.}~\cite{kan2014stacked} design stacked auto-encoders to progressive frontalize non-frontal HR faces based on the learned pose-robust features.
Cole \textit{et al.}~\cite{cole2017face} present a face recognition network to decompose an input face image into facial landmarks and aligned texture maps. 
Then, a differentiable image warping operation is conducted to produce the frontal view. 
Tuan \textit{et al.}~\cite{tuan2017regressing} propose a CNN-based model to learn 3DMM shapes as well as texture parameters.
However, they render frontal faces without taking the image intensity similarity into consideration, resulting in distorted results.

Later, image generative models~\cite{goodfellow2014generative} have been widely employed for face frontalization.
Tran \textit{et al.}~\cite{tran2018representation} propose a GAN-based model to learn the disentangled representation of input faces, achieving label-assisted face frontalization and pose-invariant face recognition.
Jie \textit{et al.}~\cite{cao2019towards} propose a high fidelity pose invariant model to synthesize frontal faces from estimated dense correspondence fields and recovered facial texture maps.

Above all, face frontalization is treated as an HR image-to-image translation problem {\color{black}{without taking face SR into consideration simultaneously.}}
On the contrary, {\color{black}{our goal is to frontalize profile faces in low resolution as well as super-resolve facial details at the same time.}}

\section{Proposed Method: VividGAN}
\begin{figure*}[t]
\centering
\includegraphics[height=3.2cm,width=1\textwidth]{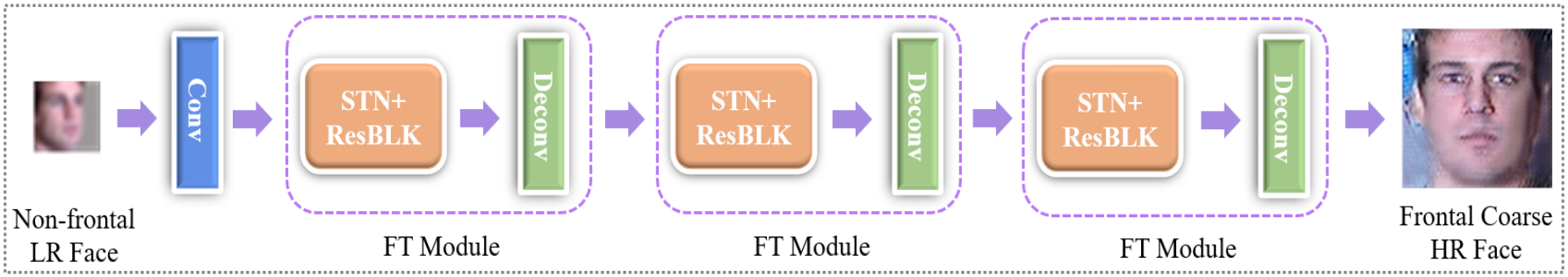}
\caption{The architecture of our coarse-level FH network.}
\label{fig3}
\vspace{-0.5cm}
\end{figure*}

Our VividGAN consists of a Vivid-FHnet that comprises a coarse-level FH network and a fine-level FH network, and two discriminators, \emph{i.e.,} Coarse-D and Fine-D.
{\color{black}{We let the coarse-level FH network provide a preview for the fine-level FH network.}}
Meanwhile, {\color{black}{two-level discriminators}} are introduced to force the hallucinated frontal HR faces to be {\color{black}{authentic as real frontal face images.}}

\subsection{Coarse-level FH Network} 
{\color{black}{Our coarse-level FH network is designed to recover a frontal coarse HR face image.}}
The architecture of our coarse-level FH network is illustrated in Fig.~\ref{fig3}. 
{\color{black}{An input LR face is firstly encoded to latent features.}} 
Then, we hallucinate the latent features by a cascade of {\color{black}{Feature Transform modules (FT module) composed of}} a spatial transform network (STN)~\cite{jaderberg2015spatial}, a residual block~\cite{he2016deep}, and a deconvolutional layer. 
Inspired by~\cite{yu2019can,yu2019hallucinating}, in each FT module, the spatial transform network (STN)~\cite{jaderberg2015spatial} layer is used to line up the intermediate features.
Afterwards, the deconvolutional layer is adopted to upsample the aligned features.
Meanwhile, the residual block~\cite{he2016deep} is introduced to {\color{black}{improve the recovery of high-frequency details and the network capacity.}}
Since the STN and upsampling layers are interwoven together, our coarse-level FH network can effectively {\color{black}{eliminate misalignments.}}

{\color{black}{To minimize discrepancies between output images and ground-truth images, our network is trained using two losses: the first one is a pixel-wise $l_2$ loss to maintain the pixel-wise intensity similarity and the second one is a perceptual loss~\cite{johnson2016perceptual} to enforce the feature-wise high-level similarity. To cope with the self-occlusion and preserve the content-integrity in output images, we introduce a mirror symmetry loss.
To enforce the output images to resemble real ones, a discriminative loss is also employed.}}
{\color{black}{As shown in Fig.~\ref{figloss}(b), our coarse-level FH network effectively performs face SR and frontalization on the unaligned non-frontal LR face, which provides a more precise facial skeleton for subsequent touch-ups.}}

\subsection{Fine-level FH Network}
{\color{black}{Our fine-level FH network is proposed to recover facial fine details of HR frontal faces by explicitly exploiting the facial component appearance prior, \emph{i.e.,} detailed facial components.
First, we design a touching-up subnetwork to recover the fine details of each facial component.
The recovered facial components will be exploited as priors to recover high-quality frontal face images.
Then, we develop a fine-integration subnetwork to merge the frontal coarse HR face and the facial component appearance prior in a seamless fashion.}}

\subsubsection{Touching-up subnetwork}
\begin{figure}[t]
\centering
\begin{minipage}[b]{.49\textwidth}
\centering
\includegraphics[height=3.2cm,width=0.95\linewidth]{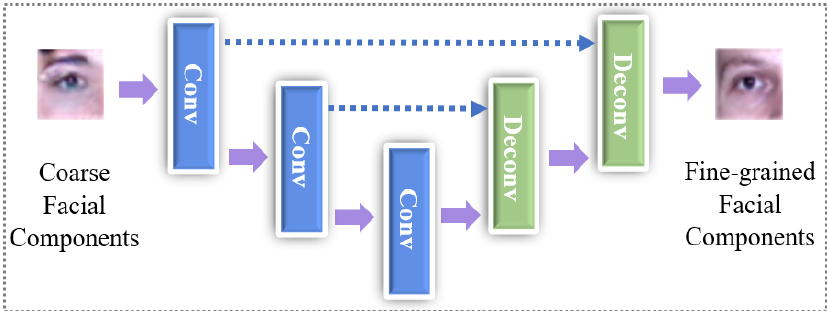}
\subcaption{{\color{black}{The architecture of the touching-up subnetwork}}}
\vspace{0.1cm}
\end{minipage}
\begin{minipage}[b]{.49\textwidth}
\centering
\includegraphics[height=3.6cm,width=0.95\textwidth]{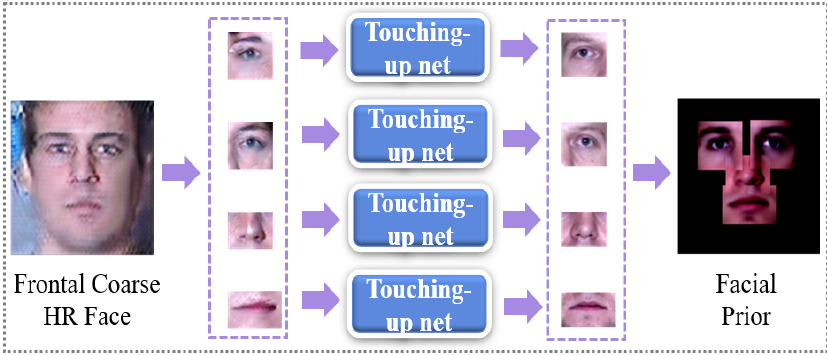}
\subcaption{{\color{black}{The generation procedure of the facial component appearance prior}}}
\end{minipage}
\caption{{\color{black}{The detail of our touching-up subnetwork.}}}
\vspace{-0.5cm}
\label{fig4}
\end{figure}

{\color{black}{After obtaining the frontal coarse HR face, obscure details of the vital facial components, \emph{i.e.,} eyes, noses, mouths, become more distinguishable and can be easily cropped out based on detected facial landmarks now.
Here, the landmark localization can be obtained by a widely-used facial landmark detector~\cite{Bulat_2018_CVPR}.}}
With the cropped parts, we design a touching-up subnetwork to refine the coarsely upsampled facial components and estimate the fine details for each component (see Fig.~\ref{fig4}(a)).

Inspired by the U-net architecture~\cite{chen2018learning}, {\color{black}{our touching-up subnetwork employs successive convolutional and deconvolutional layers along with skip connections to enhance features of each component.
In this way, multi-scale features of facial components can be integrated to achieve photo-realistic results.}}

Towards this goal, {\color{black}{we establish facial component sets of the regions of left eyes, right eyes, noses and mouths}} on the Multi-PIE database~\cite{gross2010multi}, the MMI Facial Expression (MMI) database~\cite{Valstar2010idhas}, the Celebrity Face Attribute (CelebA) database~\cite{liu2015faceattributes} and our proposed PSD-HIGHROAD database~\cite{zhang2019face}, respectively, {\color{black} which covers the traits from different genders, races, ages and facial expressions.}

{\color{black}{We pre-train the touching-up subnetwork with a pixel-wise intensity similarity loss on these facial component sets for parameter initialization.
Afterwards, the estimated fine-grained facial components are stitched onto a masked template as our facial component appearance prior (see Fig.~\ref{fig4}(b)).}}
Meanwhile, the max-out fusing strategy is adopted to reduce the stitching artifacts on the overlapping areas.

\subsubsection{Fine-integration subnetwork}
\begin{figure}[t]
\centering
\begin{minipage}[b]{.48\textwidth}
\centering
\includegraphics[width=0.96\textwidth]{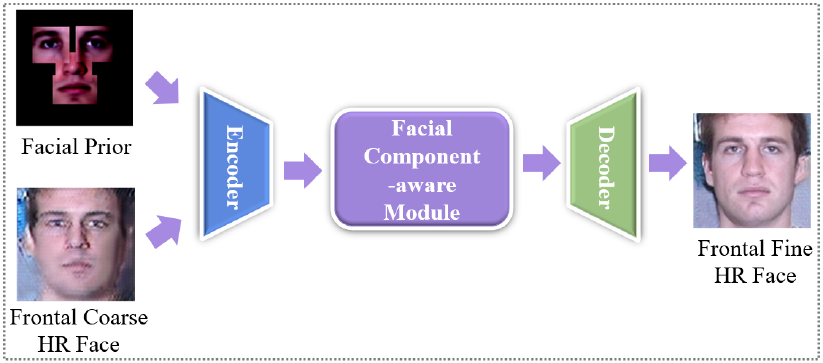}

\subcaption{The architecture of the fine-integration subnetwork}
\vspace{0.1cm}
\end{minipage}
\begin{minipage}[b]{.45\textwidth}
\centering
\includegraphics[width=0.98\linewidth]{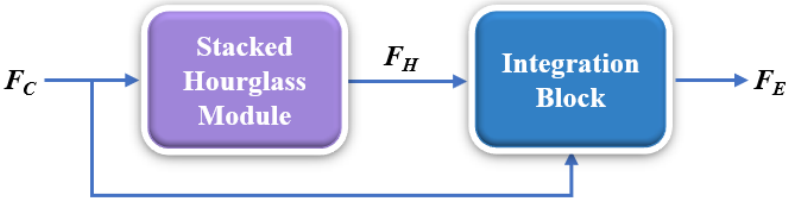}

\subcaption{The facial component-aware module}
\vspace{0.1cm}
\end{minipage}
\begin{minipage}[b]{.48\textwidth}
\centering
\includegraphics[width=0.85\linewidth]{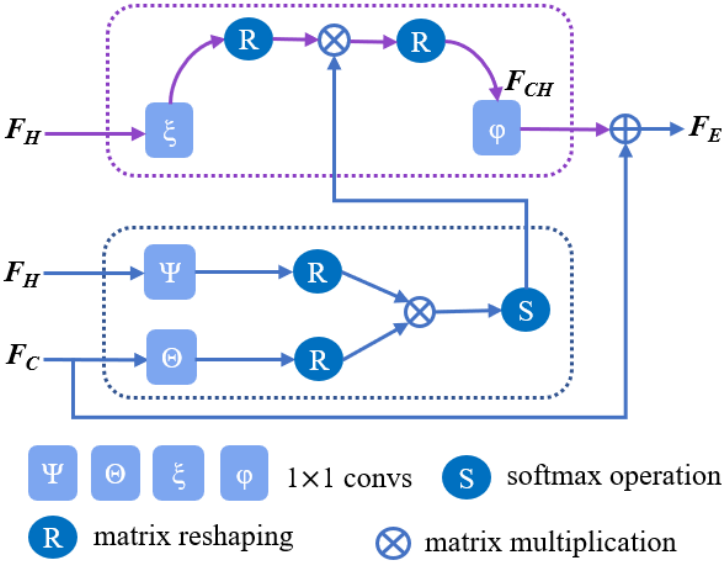}

\subcaption{The processing procedure of the integration block}
\vspace{0.1cm}
\end{minipage}
\caption{The detail of our fine-integration subnetwork.}
\vspace{-0.5cm}
\label{figure6}
\end{figure}

{\color{black}{We feed the facial component appearance prior and the frontal coarse HR face as a blueprint into a fine-integration subnetwork for further refinement.}}
Our fine-integration subnetwork generates the frontal fine HR face in three steps, as illustrated in Fig.~\ref{figure6}(a):
$\left(\romannumeral1\right)$ an \emph{encoder} concatenates the frontal coarse HR face and the component appearance prior to produce the fused features, denoted as $F_C$;
$\left(\romannumeral2\right)$ a \emph{facial component-aware module} aligns and merges the coarsely restored face and the stitched facial components in the feature level;
$\left(\romannumeral3\right)$ a \emph{decoder} reconstructs a frontal fine HR face.
{\color{black}In the fine-integration subnetwork, the encoder and decoder adopt the same architectures as our coarse-level FH network except that the encoder of the fine-integration subnetwork takes two images as input (\emph{i.e.}, a coarsely frontalized and upsampled face and a stitiched component prior image).}

{\color{black}{Our newly designed facial component-aware module consists of a staked hourglass network~\cite{newell2016stacked} and an integration block (see Fig.~\ref{figure6}(b)).}}
First, we introduce a staked hourglass network~\cite{newell2016stacked} to estimate the facial landmark heatmaps from the fused features.
Then, we feed the facial landmark heatmaps and the fused features to an integration block for re-calibration.

\begin{figure*}[htb]
\centering
\includegraphics[width=1\textwidth]{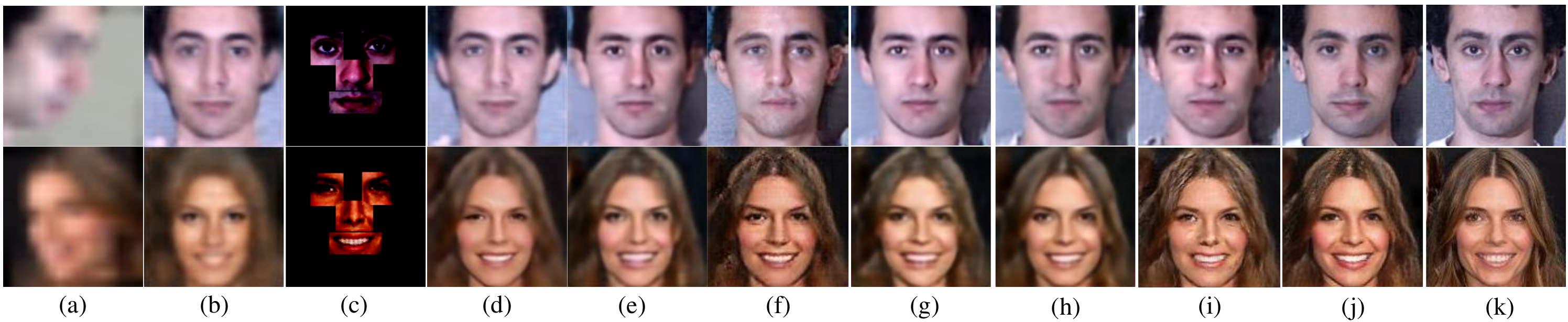}
\vspace{-0.6cm}
\caption{Ablation study on the effect of different sub-nets and losses on Multi-PIE and CelebA databases. (a) The input non-frontal LR images (16$\times$16 pixels). (b) {\color{black}{The frontal coarse HR faces.}} (c) The fine-grained facial components. (d) The results only using $L_{mse}$ and $L_{h}$. (e) The results using $L_{mse}$, $L_{id}$ and $L_{h}$. (f) The results of VividGAN w/o facial component-aware module (w/o $L_{h}$). (g) The results of Vivid-FHnet. (h) The results of Vivid-FHnet and Coarse-D. (i) The results of Vivid-FHnet and Fine-D. (j) {\color{black}{The results of VividGAN.}} (k) The ground-truth frontal HR images (128$\times$128 pixels).}
\label{figloss}
\vspace{-0.5cm}
\end{figure*}

{\color{black}{Fig.~\ref{figure6}(c) illustrates the re-calibration procedure in our integration block.}}
The facial landmark heatmaps $F_{H}$ and the fused featurs $F_C$ are first normalized and transformed into two feature spaces $\theta$ and $\psi$ to measure their similarity.
Then, the attended features $F_{CH}$ are formulated as a weighted sum of the facial landmark heatmaps $F_{H}$ that are similar to the fused features $F_{C}$ in the corresponding positions.
The $i$-th output response is expressed as:
\begin{equation}
{F_{CH}^{i}}=\frac{1}{M(F)} \sum_{\forall j} \left\{\exp \left(\boldsymbol{W}_{\theta}^{T} \boldsymbol{\left(\overline{{F_{C}^{i}}}\right)}^{T}\boldsymbol{\overline{{F_{H}^{j}}}} \boldsymbol{W}_{\psi}\right) \boldsymbol{F_{H}^{j}} \boldsymbol{W}_{\zeta}\right\},
\label{eq1}
\end{equation}
where $M(F)=\sum_{\forall j}\exp\left(\boldsymbol{W}_{\theta}^{T} \boldsymbol{\left(\overline{{F_{C}^{i}}}\right)}^{T}\boldsymbol{\overline{{F_{H}^{j}}}} \boldsymbol{W}_{\psi}\right)$ is the sum of all output responses over all positions.
In Eq.~\eqref{eq1}, the embedding transformations ${W}_{\theta}$, ${W}_{\psi}$ and ${W}_{\zeta}$ are learnt in training.

Finally, we integrate the fused features $F_{C}$ and the attended features $F_{CH}$ to form the final refined features $F_{E}$, as follows: 
\begin{equation}
F_{E}=\sigma \left(F_{CH}\boldsymbol{W}_{\varphi}\right)+F_{C},
\label{eq2}
\end{equation}
where ${W}_{\varphi}$ is also learnt during the training process. 
$\sigma$ is a trade-off parameter, and is set to 1 in our experiment.
{\color{black}{As a result, our integration block provides spatial configuration of facial components.}}

{\color{black}{During training, we employ a pixel-wise intensity similarity loss, a feature-wise identity similarity loss~\cite{johnson2016perceptual}, a structure-wise similarity loss~\cite{yu2018super} and a discriminative loss.
Once our fine-integration subnetwork is trained, it is able to achieve frontal HR faces with authentic facial details (see Fig.~\ref{figloss}(j)).}}

To verify the effectiveness of our facial component-aware module, we conduct some comparisons.
As shown in Fig.~\ref{figloss}(f), the VividGAN variant without the facial component-aware module produces inferior results.
On the contrary, our facial component-aware module achieves superior feature alignment and enhancement.

\subsection{Two-level Discriminators}
{\color{black}{Our Vivid-FHnet hallucinates non-frontal LR faces in a coarse-to-fine manner and produces high-quality HR faces. 
Inspired by~\cite{yu2019hallucinating,yu2018super}, we employ a discriminative network to force the generated faces to lie on the same manifold as real frontal faces.
However, a single discriminator commonly used in previous works~\cite{yu2019hallucinating,yu2018super,yu2017face,yu2019can,shiri2019recovering} might not be suitable for our two-level face hallucination.}}
{\color{black}{Thus, we propose two-level discriminators, \emph{i.e.,} Coarse-D and Fine-D, to address these two types of recovered HR faces.}}
The Coarse-D enforces a coarsely hallucinated face to be an upright and complete preview; while the Fine-D focuses on the fine hallucinated one for more sharper details.

{\color{black}To analyse the effect of our two-level discriminators, we perform different VividGAN variants, as shown in Figs.~\ref{figloss}(g),(h),(i) and (j).
It can be obviously seen that our VividGAN (Fig.~\ref{figloss}(j)) captures both the global outline of the face as well as detailed facial characteristics.}

\subsection{Objective Functions}
In our work, we employ five individual losses to train our networks, including a mirror symmetry loss ($L_{sys}$), a pixel-wise intensity similarity loss ($L_{mse}$), a feature-wise identity similarity loss ($L_{id}$), a structure-wise similarity loss ($L_{h}$) and a class-wise discriminative loss ($L_{adv}$).

\subsubsection{Mirror symmetry loss} 
Human faces, like many biological forms, manifest high degrees of symmetry.
{\color{black}{Thus, we introduce a mirror symmetry loss (${L}_{sys}$) to guarantee the content integrity of hallucinated faces.}} 
The mirror symmetry loss ${L}_{sys}$ of a generated image $\hat{h}_{i}$ is formulated as:
\begin{equation}
L_{sym}=\mathbb{E}_{\left(\hat{h}_{i}\right) \sim p(\hat{h})}\left\|\vec{h}_{i}-\hat{h}_{i}\right\|_{F}^{2},
\label{eqsy}
\end{equation}
where $\vec{h}_{i}$ represents the horizontally flipped copy of a generated face $\hat{h}_{i}$.

\subsubsection{Pixel-wise intensity similarity loss} 
In order to {\color{black}{force the hallucinated face to be close to its ground-truth, an intensity similarity loss ${L}_{mse}$ is employed,}} defined as:
\begin{equation}
\begin{aligned} {L}_{mse} &=\mathbb{E}_{\left(\hat{h}_{i}, h_{i}\right) \sim p(\hat{h}, h)}\left\|\hat{h}_{i}-h_{i}\right\|_{F}^{2}, \end{aligned}
\label{eqmse}
\end{equation}
where $p(\hat{h}, h)$ represents the joint distribution of the generated results $\hat{h}_{i}$ and the corresponding ground-truths $h_{i}$.

As mention in~\cite{ledig2017photo}, only employing pixel-wise intensity similarity losses ${L}_{mse}$ in training often leads to overly smoothed results and the network may fail to generate high-frequency facial features (see Fig.~\ref{figloss}(d)). Therefore, we incorporate a feature-wise identity similarity loss to enhance our hallucinated results.

\subsubsection{Feature-wise identity similarity loss}
Identity preservation is one of the most important goals in face hallucination~\cite{shiri2019identity}. 
{\color{black}{Thus, we adopt the identity similarity loss ${L}_{id}$ by measuring the Euclidean distance between the high-level features of a hallucinated face and its ground-truth, thus endowing our VividGAN with the identity preserving ability.}}
The identity similarity loss ${L}_{id}$ is expressed as:
\begin{equation}
\begin{aligned} {L}_{id} &=\mathbb{E}_{\left(\hat{h}_{i}, h_{i}\right) \sim p(\hat{h}, h)}\left\|\Phi\left(\hat{h}_{i}\right)-\Phi\left(h_{i}\right)\right\|_{F}^{2}, \end{aligned}
\label{eq3}
\end{equation}
where $\Phi(\cdot)$ represents the extracted feature vector from the average pooling layer of the Resnet50 model~\cite{he2016deep} for the input images. 
As seen in Fig.~\ref{figloss}(e), employing ${L}_{id}$ indeed improves the generated results while producing more authentic facial details.

\subsubsection{Structure-wise similarity loss}
{\color{black}{To localize spatial configuration of facial components, a structure-wise similarity loss~\cite{yu2018super} ${L}_{h}$ is employed in training our facial component-aware module,}} written as:
\begin{equation}
{L}_{h}=\mathbb{E}_{\left(l_{i}, h_{i}\right) \sim p(l, h)} \frac{1}{P} \sum_{k=1}^{P} \left\|{H}^{k}\left(f_{i}\right)-{H}^{k}\left(h_{i}\right)\right\|_{2}^{2},
\label{eq4}
\end{equation}
where ${H}^{k}\left(f_{i}\right)$ represents the $k$-th predicted facial landmark heatmap estimated from the intermediate facial features $f_{i}$ by a stacked hourglass module.
${H}^{k}\left(h_{i}\right)$ denotes the $k$-th facial landmark heatmap generated by FAN~\cite{bulat2017far} on the ground-truth image ${h}_{i}$. 
Here, we use 68 point facial landmarks to produce a heatmap.

\subsubsection{Class-wise discriminative loss}
{\color{black}{Aiming at generating photo-realistic results, we infuse the class discriminative information into our Vivid-FHnet by adopting the two-level discriminators.}}
Our goal is to let the two-level discriminators fail to distinguish hallucinated faces from ground-truth ones. 
The objective function ${L}_{D}$ for the discriminator is defined as follows:
\begin{equation}
\begin{aligned} 
{L}_{D} &=-\mathbb{E}_{\left(\hat{h}_{i}, h_{i}\right) \sim p(\hat{h}, h)}\left[\log {D}_{d}\left(h_{i}\right)+\log \left(1-{D}_{d}\left(\hat{h}_{i}\right)\right)\right],
\end{aligned}
\label{eq5}
\end{equation}
where $D$ and $d$ represent the discriminator and its parameters. 
During training, we minimize the loss ${L}_{D}$ and update the parameters for the discriminator.

{\color{black}{On the contrary, our Vivid-FHnet is designed to produce realistic face images, which would be classified as real faces by the discriminators.}}
Thus, the discriminative loss ${L}_{adv}$ is represented as:
\begin{equation}
\begin{aligned} {L}_{adv} &=-\mathbb{E}_{\hat{h}_{i} \sim p(\hat{h})} \log \left({D}_{d}\left(\hat{h}_{i}\right)\right). \end{aligned}
\label{eq6}
\end{equation}
To optimize the Vivid-FHnet, we minimize the loss ${L}_{adv}$.

\begin{figure*}[htb]
\centering
\includegraphics[height=4.5cm,width=0.98\textwidth]{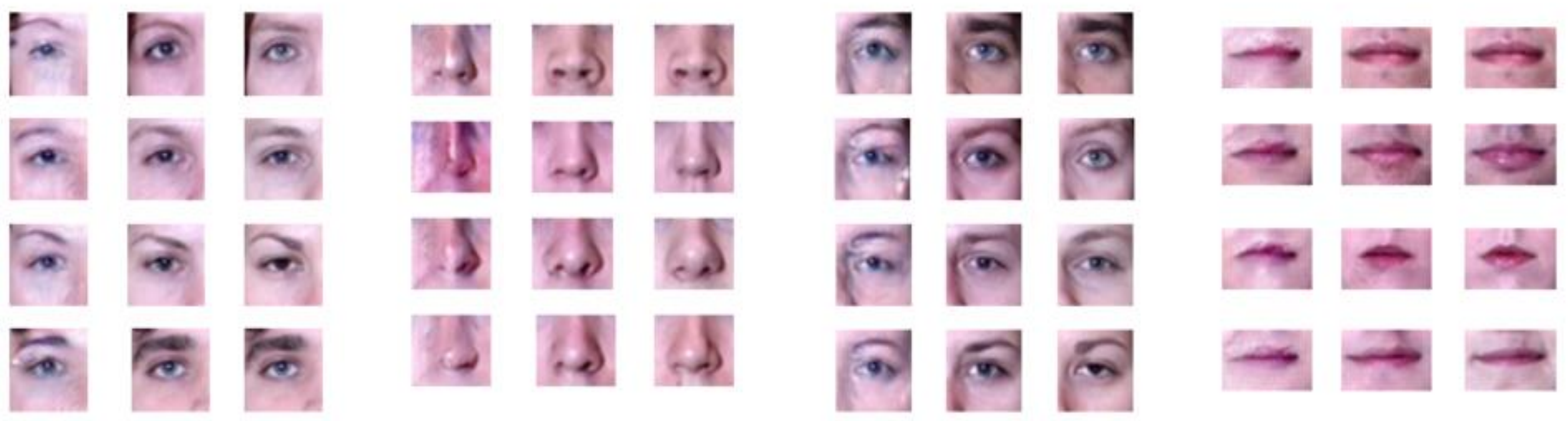}
\caption{Illustration of facial components. The coarse facial components are shown in the first column, the fine-grained facial components are illustrated in the second column, and the ground-truth facial components are presented in the third column.} 
\label{facialpart}
\vspace{-0.3cm}
\end{figure*}

\begin{figure}[htb]
\centering
\includegraphics[height=2.6cm]{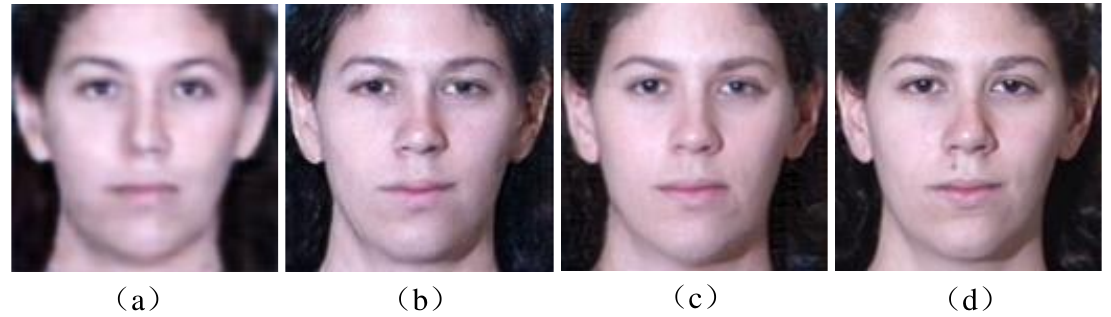}
\caption{Impacts of the facial component appearance prior. (a) Result of V-TN model (25.784/0.870 in PSNR/SSIM). (b) Result of VividGAN model (26.289/0.876 in PSNR/SSIM). (c) Result of V+GT model ({\bf 28.405/0.902} in PSNR/SSIM). (d) The ground-truth image.}
\label{figfacialparts}
\vspace{-0.4cm}
\end{figure}

\begin{figure*}[t]
\centering
\includegraphics[height=3.7cm,width=0.9\textwidth]{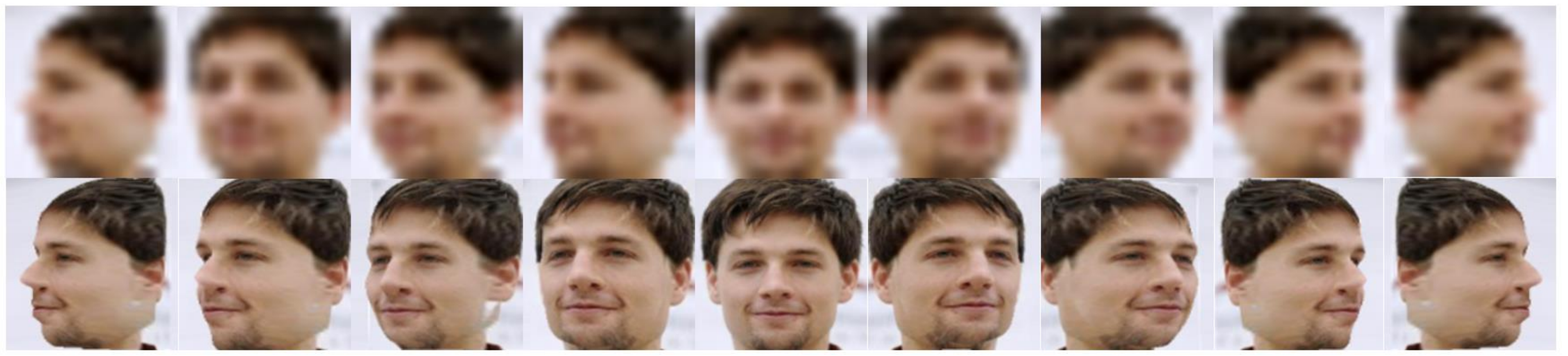}
\caption{Illustration of the synthesized non-frontal faces. First line: the synthesized non-frontal LR faces; second line: the corresponding HR ones.}
\label{fig6}
\vspace{-0.2cm}
\end{figure*}

\subsection{Training Details}
Since different subnetworks focus on different goals, we use different losses to train our coarse-level FH network, touching-up subnetwork and fine-integration subnetwork.

The objective function for the coarse-level FH network, $L_{C}$, is expressed as:
\begin{equation}
\begin{aligned} 
L_{C}= L_{mse}^{c}+L_{sys}^{c}+\alpha_{1} L_{id}^{c}+\psi_{1} {L}_{adv}^{c}.
\end{aligned}
\label{eq7}
\end{equation}

The objective function for the touching-up subnetwork, $L_{T}$, is expressed as:
\begin{equation}
\begin{aligned} 
 L_{T}= L_{mse}^{t}.
\end{aligned}
\label{eq8}
\end{equation}

The objective function for the fine-integration subnetwork, $L_{F}$, is expressed as:
\begin{equation}
\begin{aligned} 
 L_{F}= L_{mse}^{f}+\alpha_{2} L_{id}^{f}+\gamma_{2} L_{h}^{f}+\psi_{2} {L}_{adv}^{f}.
\end{aligned}
\label{eq9}
\end{equation}

Above all, the total objective function for our Vivid-FHnet, $L_{G}$, is expressed as:
\begin{equation}
\begin{aligned} 
 L_{G}= L_{C}+ L_{T}+ L_{F}.
\end{aligned}
\label{eq10}
\end{equation}

Since we intend to hallucinate frontal HR faces to be close to ground-truth appearance, we set lower weights on $L_{id}$, $L_{h}$ and $L_{adv}$.
Thus, $\alpha_{1}, \psi_{1}, \alpha_{2}, \gamma_{2}$ and $\psi_{2}$ in Eq.~\eqref{eq7}, \eqref{eq8} and \eqref{eq9} are set to 0.01. 
{\color{black}{Moreover, the training process of our VividGAN model involves in three stages:
$\left(\romannumeral1\right)$ Pre-training the coarse-level FH network with loss $L_{C}$ (Eq.~\eqref{eq7}) on the training dataset for parameter initialization;
$\left(\romannumeral2\right)$ Pre-training the touching-up subnetwork with loss $L_{T}$ (Eq.~\eqref{eq8}) on the facial component set for parameter initialization;
$\left(\romannumeral3\right)$ Training the whole VividGAN model with three losses together: Vivid-FHnet is trained by $L_{G}$ (Eq.~\eqref{eq10}) and loss ${L}_{D}$ (Eq. \ref{eq5}) is used to optimize Coarse-D and Fine-D.
In the final stage, since our fine-integration subnetwork has not been initialized, the learning rate for training our fine-integration subnetwork is set to 10$^{-3}$ while the learning rate for training the other networks is set to 10$^{-4}$.}}

\section{Prior Knowledge for Joint Face SR and Frontalization}
{\color{black}{Most of real-world objects have their distinct structured appearances, like human faces.}}
In this paper, we model and leverage {\color{black}{the facial component appearance prior}} to {\color{black}{facilitate joint face SR and frontalization.}}
As a consequence, there are two questions worthy of exploring:
$\left(\romannumeral1\right)$ Is the facial component appearance prior knowledge useful for joint face SR and frontalization? $\left(\romannumeral2\right)$ How many improvements will it bring?

{\color{black}{To answer the above questions, we conduct subjective and objective experiments on the Multi-PIE database~\cite{gross2010multi}.}}
Multi-PIE provides non-frontal/frontal face pairs of 337 individuals under various poses and illumination conditions. 
We introduce 45K face pairs, consisting of non-frontal faces ($\pm$15$^{\circ}$, $\pm$30$^{\circ}$, $\pm$45$^{\circ}$, $\pm$60$^{\circ}$, $\pm$75$^{\circ}$, $\pm$90$^{\circ}$) and corresponding frontal faces (0$^{o}$), to construct the training set.
The rest 5K face pairs are used for testing.

\subsection{Baseline Models}
We formulate two baseline models to verify that the facial component appearance prior knowledge is important for joint face SR and frontalization.

{\color{black}{The baseline models are summarized as follows (Here, we denote VividGAN as V, touching-up network as TN, and ground-truth as GT): 
 \begin{itemize}
     \item V-TN: we remove the touching-up subnetwork, and construct the ``V-TN" model. ``V-TN" model consists of the coarse-level FH network and the fine-integration subnetwork.
     \item V+GT: we introduce the ground-truth facial component appearance prior, \emph{i.e.,} ground-truth facial components, to replace the estimated prior of the touching-up subnetwork, constructing the ``V+GT" model.
 \end{itemize}
}}

\subsection{Importance of Facial Component Prior}
The results of the compared models are illustrated in Fig.~\ref{figfacialparts}.
As we can see, the results in Figs.~\ref{figfacialparts}(b) and (c) show more vivid facial details than the result in Fig.~\ref{figfacialparts}(a).
This clearly manifests the importance of the facial prior knowledge in the hallucination process.
As indicated in Fig.~\ref{figfacialparts}, V+GT model (with the ground-truth facial component appearance prior) outperforms VividGAN model (with the estimated facial component appearance prior) and V-TN model (without prior information) with the PSNR improvement of 2.116 dB and 2.621 dB, respectively.
{\color{black}{Therefore, our VididGAN provides a solution to hallucinating better fine-grained facial components, as shown in Fig.~\ref{facialpart}, and significantly improve the visual quality of the hallucinated faces.}}

\section{Experiments}
\begin{figure*}[htb]
\centering
\includegraphics[width=1\textwidth]{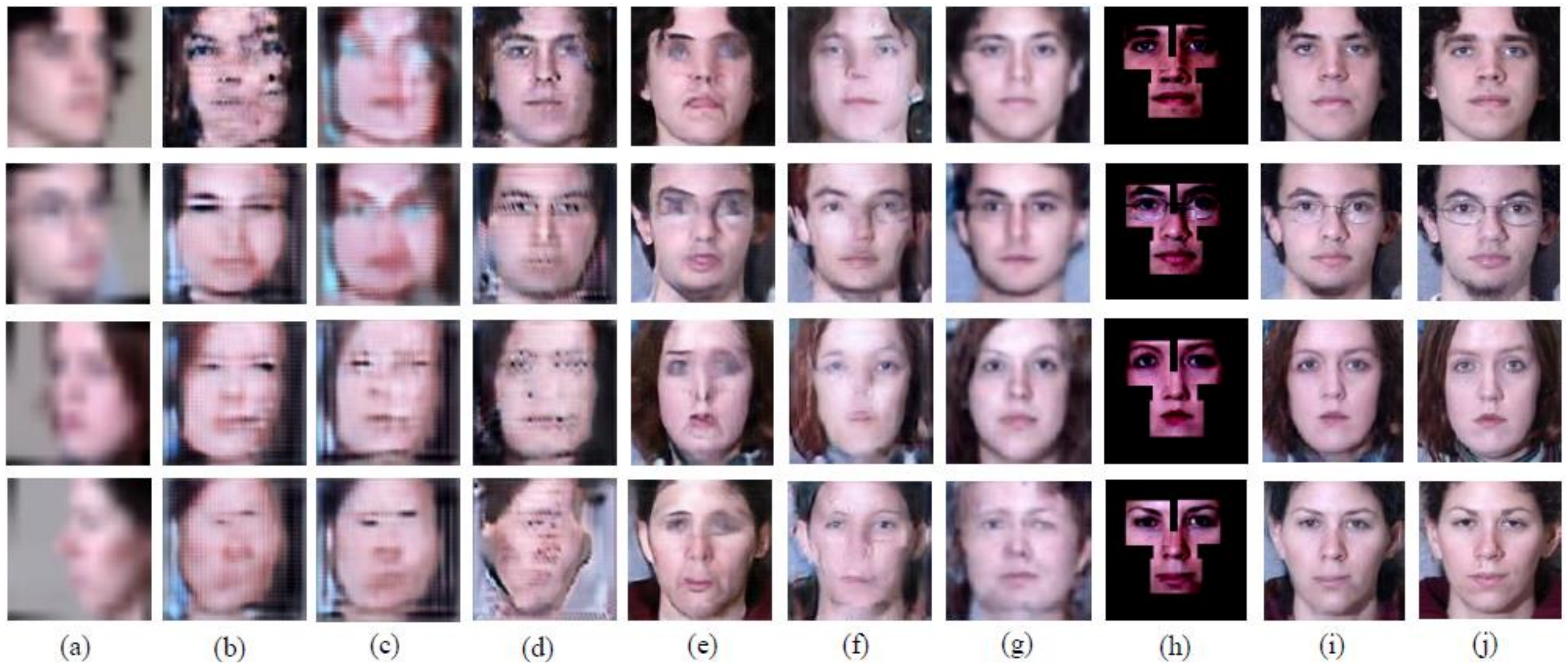}
\vspace{-0.5cm}
\caption{Qualitative comparisons of state-of-the-art methods on the Multi-PIE database. Columns: (a) Unaligned LR inputs under various poses (Rows: +60$^{o}$, +45$^{o}$, -75$^{o}$ and -90$^{o}$). (b) Bicubic +~\cite{tran2018representation} (c)~\cite{tran2018representation} +~\cite{zhu2016deep}. (d)~\cite{tran2018representation} +~\cite{huang2019wavelet}. (e)~\cite{ledig2017photo} +~\cite{tran2018representation}. (f)~\cite{yu2017face} +~\cite{tran2018representation}. (g)~\cite{yu2019can}. (h) Fine-grained facial components. (i) VividGAN. (j) Ground-truths.} 
\label{viewcompare}
\vspace{-0.5cm}
\end{figure*}

\subsection{Experimental Setup}
\subsubsection{Databases}
VividGAN is trained and tested on multiple widely used benchmarks, \emph{i.e.,} the Multi-PIE database~\cite{gross2010multi}, the MMI facial expression database~\cite{Valstar2010idhas} and the CelebA database~\cite{liu2015faceattributes}.

\textbf{Multi-PIE}~\cite{gross2010multi} has been described in Sec. \uppercase\expandafter{\romannumeral4}.
\textbf{CelebA}~\cite{liu2015faceattributes} is a large in-the-wild database that contains more than 200,000 face images under different pose, occlusion and background variations.
\textbf{MMI}~\cite{Valstar2010idhas} contains 2,900 videos and HR still images of 75 individuals.
We extract the frames of the individuals from their facial expression sequences.
Each of these frames is labeled as one of seven basic expressions, \emph{i.e.,} ``angry”, ``disgust”, ``fear”, ``happy”, ``sad”, ``surprise”, and ``neutral”.

{\color{black}{On the \textbf{Multi-PIE} database, we employ 50K face pairs consisting of non-frontal faces ($\pm$15$^{\circ}$, $\pm$30$^{\circ}$, $\pm$45$^{\circ}$, $\pm$60$^{\circ}$, $\pm$75$^{\circ}$, $\pm$90$^{\circ}$) and corresponding frontal faces (0$^{o}$), for experiments.}}
{\color{black}{\textbf{CelebA} and \textbf{MMI} databases do not provide frontal/non-frontal face pairs.}}
Thus, following \cite{yu2019can}, we use a popular 3D face model~\cite{masi2016we} to synthesize non-frontal HR face images, \emph{i.e.,} \{$\pm$22$^{\circ}$, $\pm$40$^{\circ}$, $\pm$55$^{\circ}$, $\pm$75$^{\circ}$\}, from frontal HR ones for training.
{\color{black}{Here, we first randomly select 10,000 and 900 cropped frontal HR faces from CelebA and MMI databases, resize them to $128\times 128$ pixels, and use them as our ground-truth images.}}
Then, we generate the unaligned non-frontal LR faces ($16\times 16$ pixels) by transforming and downsampling the synthesized non-frontal HR ones.
Fig.~\ref{fig6} illustrates some synthesized non-frontal faces.
{\color{black}{In this manner, we generate 80,000 and 7,200 unaligned non-frontal LR$/$frontal HR face pairs for CelebA and MMI databases.}}
{\color{black}{Finally, for each database, we choose 80 percent of the face pairs for training and 20 percent of the face pairs for testing, respectively.
In this way, the training and testing sets do not overlap.}}

\begin{figure*}[htb]
\centering
\includegraphics[width=1\textwidth]{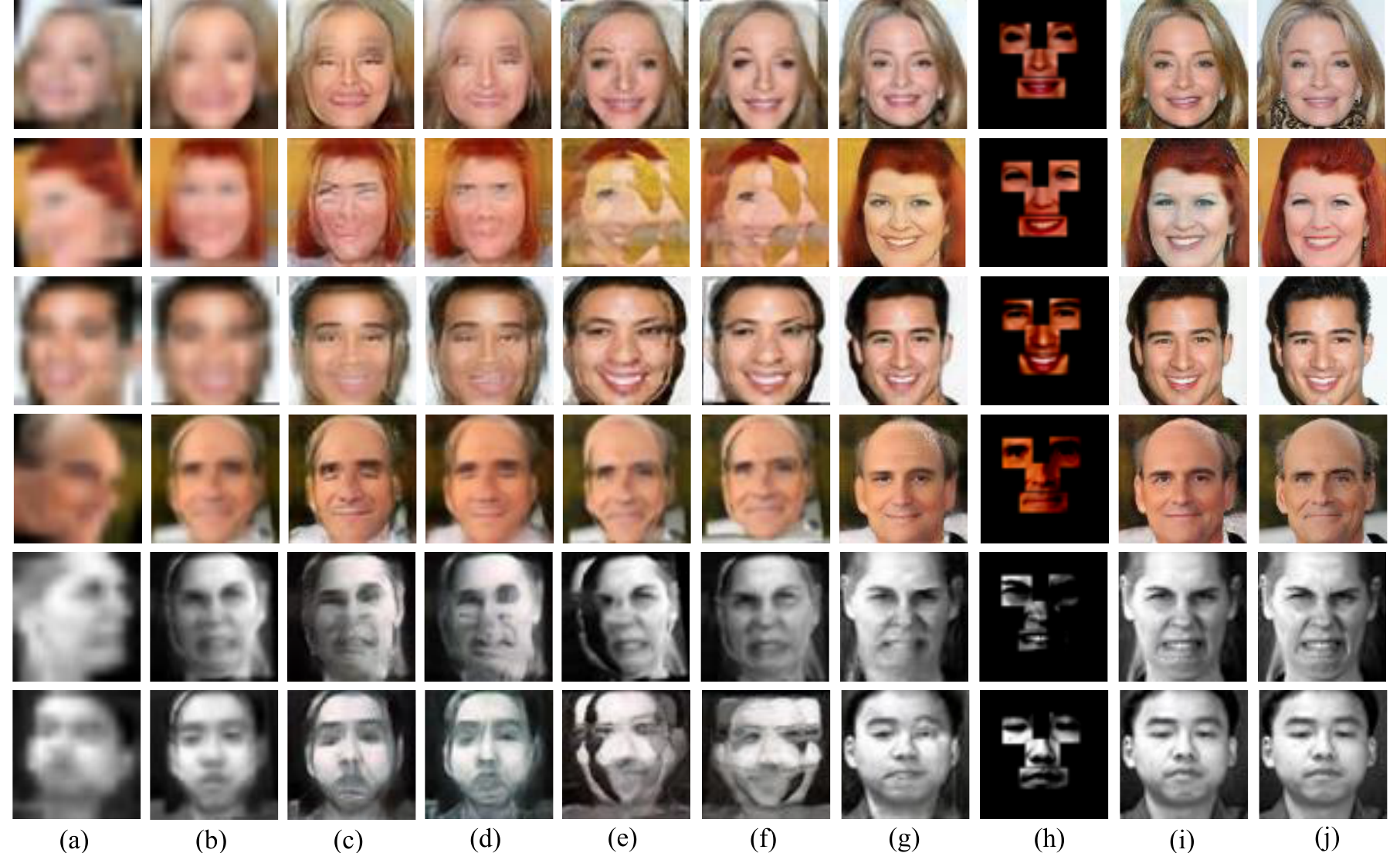}
\vspace{-0.5cm}
\caption{{\color{black}{Qualitative comparisons of state-of-the-art methods on CelebA and MMI databases. Columns: (a) Unaligned LR inputs under various poses (Rows: -40$^{o}$, -75$^{o}$, -22$^{o}$, +75$^{o}$, +75$^{o}$ and -40$^{o}$). (b) Bicubic +~\cite{hassner2015effective}. (c)~\cite{hassner2015effective} +~\cite{zhu2016deep}. (d)~\cite{hassner2015effective} +~\cite{huang2019wavelet}. (e)~\cite{ledig2017photo} +~\cite{hassner2015effective}. (f)~\cite{yu2017face} +~\cite{hassner2015effective}. (g)~\cite{yu2019can}. (h) Fine-grained facial components. (i) VividGAN. (j) Ground-truths.}}}
\label{viewcomparec}
\vspace{-0.1cm}
\end{figure*}

\begin{figure*}[htb]
\centering
\includegraphics[height=5cm,width=0.8\textwidth]{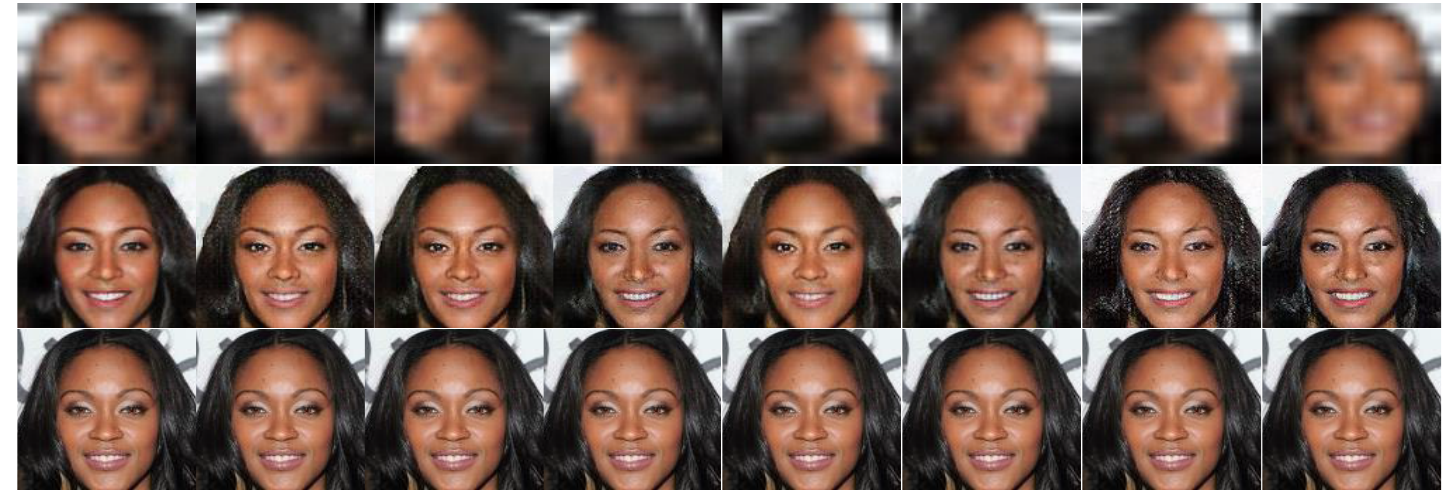}
\caption{Results of our VividGAN. First row: the input unaligned LR faces under various poses; second row: our hallucinated results; third row: the ground-truth frontal HR face images.} 
\label{expression}
\vspace{-0.5cm}
\end{figure*}

\begin{table*}[t]
\caption{{\color{black}{Average PSNR [dB] and SSIM values of the competing methods on the entire testing sets.}}}
\centering
\begin{threeparttable}
\begin{tabular}{@{}c|c|c|c|c|c|c|c|c|c|c|c|cc@{}}
\toprule
\multirow{3}{*}{SR Method} & \multicolumn{2}{c|}{Multi-PIE} & \multicolumn{2}{c|}{MMI} & \multicolumn{2}{c|}{CelebA} & \multicolumn{2}{c|}{Multi-PIE} & \multicolumn{2}{c|}{MMI} & \multicolumn{2}{c}{CelebA} \\ \cmidrule(l){2-13} 
                           & \multicolumn{6}{c|}{F~\cite{tran2018representation}+SR}                                                               & \multicolumn{6}{c}{SR+F~\cite{tran2018representation}}                                                               \\ \cmidrule(l){2-13} 
                           & PSNR           & SSIM          & PSNR        & SSIM       & PSNR          & SSIM        & PSNR           & SSIM          & PSNR        & SSIM       & PSNR          & SSIM        \\ \midrule
Bicubic       & 21.352         & 0.803         & 21.670      & 0.812      & 21.055        & 0.802                    & 18.991         & 0.730         & 19.125      & 0.738      & 20.956        & 0.799       \\ 
SRGAN~\cite{ledig2017photo}       & 21.562         & 0.804         & 21.328      & 0.809      & 20.996        & 0.800                      & 19.235         & 0.733         & 19.958      & 0.744      & 20.072        & 0.794       \\ 
CBN~\cite{zhu2016deep}               & 21.974         & 0.806         & 20.846      & 0.792      & 20.073        & 0.794               & 20.251         & 0.752         & 20.006      & 0.749      & 19.784        & 0.783        \\ 
WaveletSRnet~\cite{huang2019wavelet}      & 22.839         & 0.811         & 22.005      & 0.817      & 21.831        & 0.817                 & 21.164         & 0.798         & 20.752      & 0.762      & 21.079        & 0.805       \\ 
TDAE~\cite{yu2017face}          & 22.649         & 0.824         & 22.124      & 0.819      & 20.708        & 0.796                & 20.032         & 0.749         & 20.353      & 0.753      & 20.353        & 0.796           \\ \midrule
TANN~\cite{yu2019can}                  & {\color{black}{24.426}}         & {\color{black}{0.831}}         & {\color{black}{23.748}}      & {\color{black}{0.845}}      & {\color{black}{25.690}}        & {\color{black}{0.870}}       & {\color{black}{24.426}}          & {\color{black}{0.831}}         & {\color{black}{23.748}}      & {\color{black}{0.845}}      & {\color{black}{25.690}}        & {\color{black}{0.870}}     \\  \midrule
VividGAN$^{\dagger}$                 & 21.078         & 0.805         & 21.264      & 0.810      & 22.181        & 0.832       & 21.078          & 0.805         & 21.264      & 0.810      & 22.181        & 0.832      \\ \midrule
\textbf{VividGAN}                   & {\color{black}{\textbf{26.289}}}         & {\color{black}{\textbf{0.876}}}         & {\color{black}{\textbf{26.452}}}      & {\color{black}{\textbf{0.881}}}      & {\color{black}{\textbf{26.965}}}        & {\color{black}{\textbf{0.893}}}       & {\color{black}{\textbf{26.289}}}         & {\color{black}{\textbf{0.876}}}         & {\color{black}{\textbf{26.452}}}      & {\color{black}{\textbf{0.881}}}      & {\color{black}{\textbf{26.965}}}        & {\color{black}{\textbf{0.893}}}\\
\bottomrule
\end{tabular}
\begin{tablenotes}
\item VividGAN$^{\dagger}$ represents the coarse-level FH network in VividGAN.
\end{tablenotes} 
\end{threeparttable} 
\label{table1}
\end{table*}

\begin{table*}[t]
\caption{{\color{black}{Quantitative evaluations on different out-of-plane rotation degrees on CelebA and MMI testing sets.}}}
\begin{center}
\begin{threeparttable}
\begin{scriptsize}
\begin{tabular}{@{}c|c|c|c|c|c|c|c|c|cc@{}}
\toprule
\multirow{2}{*}{}     & \multirow{2}{*}{SR Method} & \multicolumn{2}{c|}{$\pm$22$^{\circ}$}     & \multicolumn{2}{c|}{$\pm$40$^{\circ}$}     & \multicolumn{2}{c|}{$\pm$55$^{\circ}$}     & \multicolumn{2}{c}{$\pm$75$^{\circ}$}     \\ \cmidrule(l){3-10} 
                      &                            & CelebA       & MMI          & CelebA       & MMI          & CelebA       & MMI          & CelebA       & MMI          \\ \midrule
\multirow{5}{*}{\makecell[c]{F~\cite{tran2018representation}\\+SR}} & Bicubic                    & 21.648/0.817 & 22.341/0.800 & 21.167/0.815 & 21.976/0.788 & 20.004/0.798 & 19.678/0.761 & 18.982/0.791 & 18.904/0.749 \\ 
                      & SRGAN~\cite{ledig2017photo}                      & 21.877/0.819 & 22.109/0.797 & 21.032/0.814 & 21.893/0.787 & 18.989/0.790 & 19.241/0.752 & 17.335/0.758 & 18.036/0.741 \\  
                      & CBN~\cite{zhu2016deep}                        & 21.004/0.814 & 22.843/0.806 & 20.838/0.810 & 22.132/0.798 & 19.104/0.793 & 18.947/0.749 & 17.893/0.780 & 17.405/0.733 \\  
                      & WaveletSRnet~\cite{huang2019wavelet}                & 22.951/0.829 & 24.005/0.826 & 22.004/0.822 & 23.346/0.817 & 20.752/0.808 & 19.901/0.762 & 19.004/0.792 & 18.536/0.747 \\ 
                      & TDAE~\cite{yu2017face}                       & 21.728/0.818 & 22.903/0.810 & 21.003/0.814 & 22.142/0.798 & 19.652/0.794 & 19.096/0.751 & 18.126/0.785 & 18.002/0.740 \\ \midrule
\multirow{5}{*}{\makecell[c]{SR+\\F~\cite{tran2018representation}}} & Bicubic                    & 21.049/0.816 & 20.052/0.749 & 21.040/0.815 & 19.233/0.729 & 19.901/0.795 & 17.036/0.703 & 17.452/0.760 & 16.021/0.683 \\ 
                      & SRGAN~\cite{ledig2017photo}                      & 21.080/0.816 & 20.991/0.764 & 20.562/0.809 & 20.148/0.750 & 19.073/0.790 & 18.200/0.718 & 17.066/0.756 & 17.005/0.700 \\  
                      & CBN~\cite{zhu2016deep}                        & 20.683/0.802 & 20.984/0.763 & 19.991/0.806 & 20.571/0.761 & 18.392/0.787 & 18.852/0.731 & 16.183/0.702 & 17.003/0.699 \\  
                      & WaveletSRnet~\cite{huang2019wavelet}                & 22.004/0.822 & 22.341/0.798 & 21.358/0.818 & 21.860/0.772 & 19.996/0.796 & 18.741/0.729 & 17.742/0.764 & 16.975/0.692 \\ 
                      & TDAE~\cite{yu2017face}                       & 21.758/0.819 & 21.042/0.767 & 21.041/0.815 & 20.248/0.752 & 19.425/0.792 & 18.015/0.716 & 17.002/0.755 & 16.897/0.690 \\  \midrule \multirow{2}{*}{\begin{tabular}[c]{@{}l@{}}Joint \\ SR+F\end{tabular}}
                      & TANN~\cite{yu2019can}                   & {\color{black}{26.709/0.882}} & {\color{black}{24.923/0.866}} & {\color{black}{25.910/0.874}} & {\color{black}{24.052/0.852}} & {\color{black}{25.362/0.870}}  & {\color{black}{23.104/0.841}} & {\color{black}{24.975/0.868}} & {\color{black}{22.621/0.834}}   \\ 
                      & {\color{black}{\textbf{VividGAN}}}                   & {\color{black}{\textbf{28.254/0.902}}} & {\color{black}{\textbf{27.490/0.893}}} & {\color{black}{\textbf{27.792/0.898}}} & {\color{black}{\textbf{26.601/0.884}}} & {\color{black}{\textbf{26.667/0.889}}} & {\color{black}{\textbf{25.712/0.876}}} & {\color{black}{\textbf{26.105/0.882}}} & {\color{black}{\textbf{25.039/0.869}}} \\ \bottomrule
\end{tabular}
\end{scriptsize}
\begin{tablenotes}
\item In each cell, the first and second numbers denote PSNR [dB] and SSIM values, respectively.
\end{tablenotes} 
\end{threeparttable} 
\end{center}
\label{table-roation}
\vspace{-0.5cm}
\end{table*}

\subsubsection{Competing methods}
We conduct comparative experiments in the following three fashions:
 \begin{itemize}
     \item F+SR: face frontalization techniques (DRGAN~\cite{tran2018representation} or Hassner\textit{et al.}~\cite{hassner2015effective}) followed by face SR methods (SRGAN~\cite{ledig2017photo}, CBN~\cite{zhu2016deep}, WaveletSRnet~\cite{huang2019wavelet} or TDAE~\cite{yu2017face}) (we use bicubic interpolation to adjust image sizes);
     \item SR+F: face SR methods (SRGAN~\cite{ledig2017photo}, CBN~\cite{zhu2016deep}, WaveletSRnet~\cite{huang2019wavelet} or TDAE~\cite{yu2017face}) followed by face frontalization techniques (DRGAN~\cite{tran2018representation} or Hassner\textit{et al.}~\cite{hassner2015effective}); 
      \item Joint SR+F: TANN~\cite{yu2019can} and our VividGAN.
 \end{itemize}

In the first fashion (F+SR), we first frontalize the non-frontal LR faces by popular frontalization techniques, and then super-resolve the frontalized faces by state-of-the-art SR methods.
In the second fashion (SR+F), we first super-resolve the non-frontal LR faces, and then frontalize the upsampled results.
In the third fashion (Joint SR+F), both TANN~\cite{yu2019can} and VividGAN jointly tackle face SR and face frontalization in a unified framework.

{\color{black}{For a fair comparison, we retrain these baseline methods on our training sets.
Since SRGAN~\cite{ledig2017photo}, CBN~\cite{zhu2016deep} and WaveletSRnet~\cite{huang2019wavelet} cannot achieve face alignment during their upsampling procedure, we train a STN~\cite{jaderberg2015spatial} to align the input unaligned LR faces to the upright position firstly.
However, TANN~\cite{yu2019can}, TDAE~\cite{yu2017face} and our VividGAN do not need any alignment in advance and generate upright results automatically.
Furthermore, compared to DRGAN~\cite{tran2018representation} and Hassner\textit{et al.}~\cite{hassner2015effective}, our VividGAN does not need auxiliary poses or facial landmarks for face frontalization.}}

\subsection{Qualitative Evaluation}
{\color{black}{Figs.~\ref{viewcompare} and \ref{viewcomparec} illustrate the visual results of the compared methods. 
The hallucinated results obtained by VividGAN are more photo-realistic and identity-preserving.}}

As shown in Figs.~\ref{viewcompare}(b) and \ref{viewcomparec}(b), the combination of bicubic interpolation and frontalization methods~\cite{tran2018representation,hassner2015effective} fails to generate photo-realistic facial details.
Since bicubic upsampling only interpolates new pixels from neighboring pixels without generating new contents, the produced non-frontal HR images lack details.
Thus, the face frontalization method fails to detect facial landmarks and outputs erroneous frontalized faces with severe artifacts and distorted contours.

{\color{black}{As discussed in Sec.~\uppercase\expandafter{\romannumeral1}, simply combining existing face SR and frontalization methods cannot address this challenging issue.}}
This is verified by the results of the F+SR methods (see Figs.~\ref{viewcompare}(c), \ref{viewcompare}(d), \ref{viewcomparec}(c) and \ref{viewcomparec}(d)), where upsampled face regions suffer severe distortions and ghosting artifacts.
Similarly, the SR+F methods also fail to recover authentic facial details (see Figs.~\ref{viewcompare}(e), \ref{viewcompare}(f), \ref{viewcomparec}(e) and \ref{viewcomparec}(f)).

{\color{black}{TANN~\cite{yu2019can} is the first attempt to jointly address face SR and face frontalization in a whole framework.
In this manner, two tasks alternately facilitate each other.
Thus, TANN generates satisfying results, as shown in Figs.~\ref{viewcompare}(g) and \ref{viewcomparec}(g).
However, due to its single-stage mechanism, TANN does not have a ``looking back” mechanism to revise the hallucinated faces.
Therefore, when input LR faces undergo extreme poses (see the forth row in Fig.~\ref{viewcompare}(a)) or complex expressions (see the fifth and sixth rows in Fig.~\ref{viewcomparec}(a)), TANN produces distorted facial details.
Moreover, the hallucinated faces suffer mild blurs, as visible in Fig.~\ref{viewcompare}(g) and Fig.~\ref{viewcomparec}(g).
}}

{\color{black}{Our VividGAN generates photo-realistic frontal HR faces from very LR inputs with various poses and expressions in Figs.~\ref{viewcompare}(i), \ref{viewcomparec}(i) and \ref{expression}.
Moreover, VividGAN is able to revise facial details and thus obtains visually appealing HR face images (\emph{e.g.,} the mouth in the fifth row of Fig.~\ref{viewcomparec}(i) and the ``slight closed eyes'' in the sixth row of Fig.~\ref{viewcomparec}(i)).
Since our VividGAN hallucinates non-frontal LR faces in a coarse-to-fine manner and embodies a revision mechanism, it reconstructs fine facial details and achieves superior hallucination performance.
}}

\subsection{Quantitative Evaluation}
Following the quantitative evaluation in TANN, we also report average PSNR and SSIM values on the testing sets in two different manners (see Tabs.~\ref{table1} and \ref{table-roation}).

As shown in Tab.~\ref{table1}, our VividGAN achieves remarkable quantitative results than other state-of-the-art methods on both under-controlled and in-the-wild databases.
Specifically, on the MMI testing set, VividGAN outperforms the second best technique TANN with a large margin of approximate 2.7 dB in PSNR.
{\color{black}{This is mainly due to our fine-level FH network that not only exploits facial component priors but also processes a revision mechanism.
Thus, the hallucinated faces by our VividGAN are more similar to the ground-truths.}}

As presented in Tab.~\ref{table-roation}, VividGAN achieves the best performance across all poses, especially for large yaw angles.
{\color{black}{Moreover, as the rotation angle increases, our method does not degrade like the other methods.
In the extremely challenging angles ($\pm$75$^{\circ}$), VividGAN outperforms the second best TANN more than 1 dB in PSNR.
This also verifies that VividGAN attains more authentic results when the input LR faces are under extreme poses.}}

\begin{table}[t]
\caption{Ablation study of different losses}
\centering
\begin{tabular}{@{}c|c|c|c|cc@{}}
\toprule
\multirow{2}{*}{} & \multicolumn{2}{c|}{Multi-PIE} & \multicolumn{2}{c}{CelebA} \\ \cmidrule(l){2-5} 
                                  & PSNR           & SSIM         & PSNR         & SSIM        \\ \midrule
$L_{G}^{-}$                         & 23.017         & 0.779        & 22.945       & 0.773       \\
$L_{G}^{\dagger}$                      & 24.332         & 0.803        & 23.981       & 0.798       \\
$L_{G}^{\ddagger}$                    & 25.998         & 0.845        & 25.057       & 0.804       \\
$L_{G}^{\star}$                & 26.055         & 0.874        & 26.128       & 0.885       \\
$L_{G}$            & \textbf{26.289}         & \textbf{0.876}        & \textbf{26.965}       & \textbf{0.893}       \\ \bottomrule
\end{tabular}
\label{table2}
\vspace{-0.4cm}
\end{table}

\begin{figure}[t]
\centering
\includegraphics[height=3.8cm,width=0.48\textwidth]{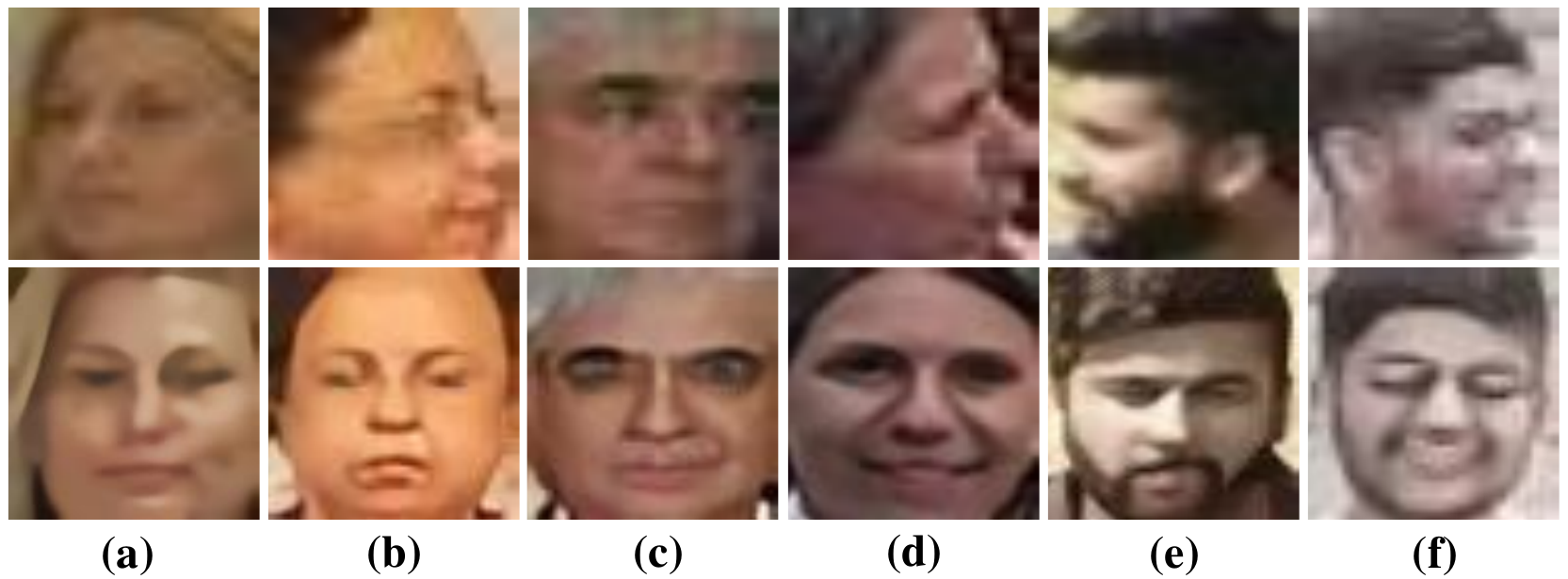}
\vspace{-0.1cm}
\caption{\color{black}{Results on LR face images in the wild. Top row: LR images in the wild. Bottom row: our hallucinated results.}}
\vspace{-0.1cm}
\label{figreal}
\end{figure}

\subsection{Ablation Analysis}
We report the performance of different VividGAN variants, which are trained with different loss combinations, on Multi-PIE and CelebA (see Tab.~\ref{table2} and Fig.~\ref{figloss}).
We denote the compared VividGAN variants as follows: $\left(\romannumeral1\right)$ $L_{G}^{-}$: $L_{mse}$, $L_{h}$; $\left(\romannumeral2\right)$ $L_{G}^{\dagger}$: $L_{mse}$, $L_{id}$ and $L_{h}$; $\left(\romannumeral3\right)$ $L_{G}^{\ddagger}$: $L_{mse}$, $L_{id}$, $L_{h}$ and $L_{sys}$; $\left(\romannumeral4\right)$ $L_{G}^{\star}$: $L_{mse}$, $L_{id}$, $L_{h}$, $L_{sys}$ and $L_{adv}^{c}$; $\left(\romannumeral5\right)$ $L_{G}$: $L_{mse}$, $L_{id}$, $L_{h}$, $L_{sys}$, $L_{adv}^{c}$ and $L_{adv}^{f}$. 
Note that $L_{h}$ is a prerequisite constraint in training our facial component-aware module.

As demonstrated in Tab.~\ref{table2}, only adopting intensity similarity loss ${L}_{mse}$ leads to unpleasant quantitative results ($L_{G}^{-}$ in Tab.~\ref{table2}).
{\color{black}{The identity similarity loss $L_{id}$ not only improves the visual quality (see Fig.~\ref{figloss}(e)) but also increases the quantitative performance ($L_{G}^{\dagger}$ in Tab.~\ref{table2}).
This experiment asserts that $L_{id}$ forces the high-order moments of the hallucinated faces, \emph{i.e.,} feature maps, to be similar to their ground-truths and thus improves hallucination performance.
In addition, we also verify the effectiveness of the mirror symmetry loss $L_{sys}$.
As indicated in Tab.~\ref{table2} ($L_{G}^{\ddagger}$), using $L_{sys}$ leads to better quantitative performance.
Since $L_{sys}$ enforces the content integrity of frontalized faces, it is able to reduce the reconstruction errors of coarsely frontal HR faces in the coarse-level FH network rather than spreading the errors through the entire Vivid-FHnet.
Thus, the fine-level FH network can focus on learning mappings between coarse and fine HR facial patterns.
Moreover, as demonstrated in Tab.~\ref{table2} ($L_{adv}^{c}$) and Fig.~\ref{figloss}(h), the Coarse-D improves the quantitative performance by forcing the coarsely hallucinated faces to be frontal and aligned.
However, Fig.~\ref{figloss}(h) shows that the hallucinated faces still suffer from mild blurriness.
Finally, the help of the two-level discriminators, VividGAN achieves photo-realistic images (see Fig.~\ref{figloss}(j)) and the best performance ($L_{G}$ in Tab.~\ref{table2}).}}

\begin{table}[t]
\centering
\caption{{\color{black}{Efficiency comparison on the CelebA testing set.}}}
\begin{threeparttable}
\begin{tabular}{@{}c|c|c|c|c@{}}
\toprule
\multicolumn{2}{c|}{Method}                                                         & \begin{tabular}[c]{@{}c@{}}Parameter\\ Number\\ (KB)\end{tabular} & \begin{tabular}[c]{@{}c@{}}Running\\ Time\\ (ms)\end{tabular} & PSNR (dB)     \\ \midrule
\multirow{4}{*}{\begin{tabular}[c]{@{}c@{}}F+SR/\\ SR+F\end{tabular}} & SRGAN~\cite{ledig2017photo}        & 278,758                                                           & 30.41                                                               & 20.072/20.996 \\ 
                                                                      & CBN~\cite{zhu2016deep}          & 270,562                                                           & 2028.23                                                             & 19.784/20.073 \\
                                                                      & WaveletSRnet~\cite{huang2019wavelet} & 324,458                                                           & 31.68                                                               & 21.097/21.831 \\ 
                                                                      & TDAE~\cite{yu2017face}         & 286,293                                                           & 50.42                                                               & 20.353/20.708 \\ \midrule
\multirow{3}{*}{\begin{tabular}[c]{@{}c@{}}Joint\\ SR+F\end{tabular}} & TANN~\cite{yu2019can}         & \textbf{196.231}                                                           & \textbf{15.60}                                                               & 25.690        \\ 
                                                                      & VividGAN$^{\dagger}$     & 103,473                                                           & 13.12                                                               & 22.181        \\  
                                                                      & VividGAN     & 242,089                                                           & 19.25                                                               & \textbf{26.965}        \\ \bottomrule
\end{tabular}
\begin{tablenotes}
\item VividGAN$^{\dagger}$ represents the coarse-level FH network in VividGAN.
\end{tablenotes} 
\end{threeparttable} 
\label{efficiency}
\vspace{-0.2cm}
\end{table}

\begin{table}[t]
\caption{{\color{black}{Data partition and statistics of the databases in our face recognition experiments.}}}
\centering
\begin{threeparttable}
\begin{tabular}{@{}c|c|c|c|c|c@{}}
\toprule
\multirow{2}{*}{Database} & \multicolumn{2}{c|}{Whole Dataset} & \multicolumn{2}{c|}{Testing Dataset} & \multirow{2}{*}{C/U} \\ \cmidrule(lr){2-5}
                          & Identities  & Images & Gallery         & Probe          &                      \\ \midrule
Multi-PIE~\cite{gross2010multi}                 & 337           & 157,617  & 50              & 5,000          & C                    \\ \midrule
MMI~\cite{Valstar2010idhas}                       & 75            & 7,200    & 16              & 896          & C                    \\ \midrule
CelebA~\cite{liu2015faceattributes}                    & 10,177        & 202,599  & 1,000           & 8,000         & U                    \\ \midrule
TinyFace~\cite{cheng2018low}                  & 5,139          & 169,403    & 157,871           & 3728        & U                    \\ \midrule
DroneSURF~\cite{kalra2019dronesurf}                 & 58            & 786,813  & 24              & 48,000         & U                    \\ \midrule
IJB-C~\cite{maze2018iarpa}                     & 3,531         & 148,800   & \multicolumn{2}{c|}{19,593 templates}           & U                    \\ \bottomrule
\end{tabular}
\begin{tablenotes}
\item `C’ means under-controlled, and `U’ means unconstrained.
\end{tablenotes} 
\end{threeparttable} 
\label{dataset}
\vspace{-0.3cm}
\end{table}

\subsection{Performance on in-the-wild Non-frontal LR Faces}
{\color{black}{To demonstrate that our method can be effectively generalized to in-the-wild non-frontal LR faces, we randomly select face images from real-world unconstrained databases (\emph{e.g.,} TinyFace~\cite{cheng2018low}, DroneSURF~\cite{kalra2019dronesurf}) for evaluation.
The in-the-wild LR face images are, in general, much blurrier than our training samples.
Here, we train the VividGAN model on the CelebA training set and our proposed facial component sets.
As shown in Fig.~\ref{figreal}, our VividGAN achieves pleasant frontalization and hallucination performance on such challenging images, demonstrating that it is not restricted to certain poses and under-controlled scenarios.
Four reasons account for this phenomenon:
\begin{enumerate}
\item Our coarse-level FH network super-resolves and frontalizes the input non-frontal LR face roughly, which provides a more precise facial skeleton for subsequent touch-ups. On top of this face blueprint, our fine-level FH network is tailored-designed to recover fine details of frontal HR face image by explicitly incorporating the facial component appearance prior, \emph{i.e.,} fine-grained facial components. These networks seamlessly work together to reduce the differences of 3D models and poses between the training and testing LR faces.
\item We have established the facial component sets (see Sec.~\uppercase\expandafter{\romannumeral3}. B) of vital facial components from the in-the-wild database, which cover the statistics from a wide range of faces.
Our touching-up subnetwork is fed with these facial component sets to provide an informative prior of vivid facial components. Thus, it can effectively hallucinate the fine-grained facial components in real-world scenarios.
\item Our proposed facial component-aware module leverages the facial landmark heatmaps as queries to accurately locate the corresponding vital facial parts. This module not only facilitates alignment but also provides spatial configuration of facial components, such as the shape of mouth. Thus, our method is able to produce visually pleasing HR facial details.
\item When generating our training set, the selected faces used for generating non-frontal faces are not required to be exactly frontal ones, which increases the variety of the training poses and thus provides flexibility to handle in-the-wild LR faces.
\end{enumerate}
}}

\begin{table*}[htb]
\caption{{\color{black}{Face recognition performance comparison on under-controlled and in-the-wild databases.}}}
\centering
\begin{threeparttable}
\begin{tabular}{@{}c|cccc|cccc|cccccccccccccccc@{}}
\toprule
Database   & \multicolumn{4}{c|}{Multi-PIE~\cite{gross2010multi}}                               & \multicolumn{4}{c|}{CelebA~\cite{liu2015faceattributes}}                                  & \multicolumn{4}{c}{MMI~\cite{Valstar2010idhas}}                                     \\  \cmidrule(l){1-1} \cmidrule(l){2-5} \cmidrule(l){6-9} \cmidrule(l){10-13}
Settings  & \multicolumn{2}{c}{$32\times 32$, $4\times$} & \multicolumn{2}{c|}{$16\times 16$, $8\times$} & \multicolumn{2}{c}{$32\times 32$, $4\times$} & \multicolumn{2}{c|}{$16\times 16$, $8\times$} & \multicolumn{2}{c}{$32\times 32$, $4\times$} & \multicolumn{2}{c}{$16\times 16$, $8\times$} \\ \cmidrule(l){1-1} \cmidrule(l){2-3}  \cmidrule(l){4-5} \cmidrule(l){6-7} \cmidrule(l){8-9}  \cmidrule(l){10-11} \cmidrule(l){12-13} 
Metric    & Rank-1        & Rank-5       & Rank-1        & Rank-5       & Rank-1        & Rank-5       & Rank-1        & Rank-5       & Rank-1        & Rank-5       & Rank-1        & Rank-5       \\ \cmidrule(l){1-1} \cmidrule(l){2-2}  \cmidrule(l){3-3}  \cmidrule(l){4-4} \cmidrule(l){5-5}  \cmidrule(l){6-6} \cmidrule(l){7-7}  \cmidrule(l){8-8}  \cmidrule(l){9-9}  \cmidrule(l){10-10} \cmidrule(l){11-11}  \cmidrule(l){12-12} \cmidrule(l){13-13}
LR face   &  51.08\%             &   62.34\%           & 40.26\%       &   48.70\%            & 3.85\%              &  9.60\%            & 1.75\%              & 6.15\%       &  11.28\%             & 18.42\%             &  8.71\%             & 13.73\%             \\
TANN~\cite{yu2019can}      & 74.04\%              & 80.32\%             & 68.50\%       &  73.16\%            & 83.40\%              &  89.55\%            &  81.35\%             & 86.70\%      &  68.30\%              &  79.13\%             &  59.71\%             &  67.97\%            \\
Vivid-GAN & \textbf{85.36\%}              & \textbf{88.04\%}             & \textbf{82.28\%}       &  \textbf{84.64\%}            & \textbf{87.60\%}              & \textbf{91.85\%}             &  \textbf{84.20\%}             & \textbf{89.30\%}      & \textbf{79.13\%}               & \textbf{90.74\%}              & \textbf{70.20\%}               &  \textbf{84.15\%}            \\ \midrule
Database   & \multicolumn{4}{c|}{TinyFace~\cite{cheng2018low}}                                & \multicolumn{4}{c|}{DroneSURF~\cite{kalra2019dronesurf}}                               & \multicolumn{4}{c}{IJB-C~\cite{maze2018iarpa}}                                   \\ \cmidrule(l){1-1} \cmidrule(l){2-5} \cmidrule(l){6-9} \cmidrule(l){10-13}
Settings  & \multicolumn{2}{c}{$32\times 32$, $4\times$} & \multicolumn{2}{c|}{$16\times 16$, $8\times$} & \multicolumn{2}{c}{$32\times 32$, $4\times$} & \multicolumn{2}{c}{$16\times 16$, $8\times$} & \multicolumn{2}{c}{$32\times 32$, $4\times$} & \multicolumn{2}{c}{$16\times 16$, $8\times$} \\  \cmidrule(l){1-1} \cmidrule(l){2-3}  \cmidrule(l){4-5} \cmidrule(l){6-7} \cmidrule(l){8-9}  \cmidrule(l){10-11} \cmidrule(l){12-13} 
Metric    & Rank-1        & Rank-5       & Rank-1        & Rank-5       & Rank-1        & Rank-5       & Rank-1        & Rank-5       & Rank-1        & Rank-5       & Rank-1        & Rank-5       \\ \cmidrule(l){1-1} \cmidrule(l){2-2}  \cmidrule(l){3-3}  \cmidrule(l){4-4} \cmidrule(l){5-5}  \cmidrule(l){6-6} \cmidrule(l){7-7}  \cmidrule(l){8-8}  \cmidrule(l){9-9}  \cmidrule(l){10-10} \cmidrule(l){11-11}  \cmidrule(l){12-12} \cmidrule(l){13-13}
LR face   &   22.96\%             &   25.70\%            & 22.96\%       & 25.70\%             &  15.82\%             &  27.07\%          & 15.82\%       & 27.07\%             &  70.09\%             &  76.33\%             & 57.26\%             &   63.17\%           \\
TANN~\cite{yu2019can}      & 35.72\%              &  40.24\%            &   30.55\%            &  34.09\%             &  18.43\%             &  32.01\%            &  17.57\%              &  30.61\%            &  81.32\%            &   85.59\%           &  78.94\%             & 82.60\%            \\
Vivid-GAN &  \textbf{47.16\%}             &  \textbf{56.04\%}            &  \textbf{45.57\%}             & \textbf{51.82\%}             & \textbf{19.65\%}              &  \textbf{35.42\%}            &  \textbf{18.54\%}             &  \textbf{33.40\%}            &  \textbf{85.65\%}              & \textbf{90.13\%}              &  \textbf{82.18\%}             & \textbf{86.32\%}              \\ \bottomrule
\end{tabular}
\begin{tablenotes}
\item ``$32\times 32$, $4\times$": $32\times 32$ means the resolution of the original non-frontal LR face; $4\times$ means the magnification factor.
\end{tablenotes} 
\end{threeparttable} 
\label{tablefr}
\vspace{-0.3cm}
\end{table*}

\subsection{Efficiency Analysis}
{\color{black}{We also conduct comparisons to verify the efficiency of our model.
As indicated in Tab.~\ref{efficiency}, compared with ``F+SR" and ``SR+F" methods, our Vivid-GAN requires relatively shorter running time because VividGAN performs face SR and frontalization seamlessly.
Here, ``F+SR" and ``SR+F" methods adopt the work~\cite{tran2018representation} for face frontalization.
Due to the coarse-to-fine mechanism, VividGAN is less efficient than TANN, and the parameter capacity of VividGAN is also slightly overweight than TANN.
However, we achieve much better improvements on quantitative performance, as indicated in Tab.~\ref{efficiency}.
As obtaining more authentic HR faces is more important for face hallucination approaches, the trade-off between hallucination performance and model complexity in our VividGAN is acceptable.}}

\section{Face Hallucination Evaluation via Downstream Tasks}
\subsection{Comparisons with SOTA on Face Recognition}
{\color{black}{First of all, we demonstrate that our VividGAN boosts the performance of low resolution face recognition.
We adopt the ``recognition via hallucination” framework to conduct face recognition experiments on both under-controlled and unconstrained databases (see Tab.~\ref{dataset}).
Concretely, aggressively downsampled or in-the-wild LR faces are first hallucinated by face hallucination methods and then used for recognition.

\subsubsection{Databases}
We partition \textbf{Multi-PIE}~\cite{gross2010multi}, \textbf{MMI}~\cite{Valstar2010idhas} and \textbf{CelebA}~\cite{liu2015faceattributes} databases into subject disjoint training and testing sets.
We also train the compared face hallucination methods and then conduct face recognition experiments on the testing set.
Tab.~\ref{dataset} manifests the detailed structure as well as the experimental protocol for each database.
In our face recognition experiments, one frontal image with natural illumination and neutral expression is used as the gallery image for each testing identity.
Meanwhile, the probe set is made of non-frontal LR face images.

For the in-the-wild setting, we train the face hallucination methods on CelebA, and then test on TinyFace~\cite{cheng2018low}, DroneSURF~\cite{cheng2018low} and IJB-C~\cite{maze2018iarpa} databases.
\textbf{TinyFace}~\cite{cheng2018low} is a face database established specifically for native low resolution face recognition tasks. 
We use their evaluation protocol and released testing set.
\textbf{DroneSURF} is the first database proposed for research of drone based face recognition.
Following the original paper~\cite{cheng2018low}, we adopt the frame-wise identification protocol, and use the alternative frame selection technique to construct the testing set.
Here, we select the frames in surveillance scenarios.
\textbf{IARPA Janus Benchmark-C face challenge (IJB-C)}~\cite{maze2018iarpa} 
is a challenging in-the-wild face recognition database.
We employ the 1:N identification protocol, and follow the predefined multi-image face templates to generate LR probe$/$HR gallery face pairs.
Thus, our testing setting is more challenging.


\subsubsection{Experimental Settings}

For all the testing images, we crop the face regions, resize them to $128\times 128$ pixels, and thus generate our HR face images.
For Multi-PIE, MMI, CelebA and IJB-C databases, we generate the non-frontal LR faces ($16\times 16$ or $32\times 32$ pixels) by transforming and downsampling the non-frontal HR ones.
As the face images in TinyFace and DroneSURF databases are captured in real-life low resolution condition, we employ them as LR images directly.
Then, we employ a state-of-the-art pretrained face recognition model (SphereFaceNet~\cite{Liu_2017_CVPR}) to conduct face recognition experiments on LR faces and hallucinated HR faces from LR ones by different methods.
We compute the cosine distance of the extracted deep features for face recognition.
In particular, we train a CycleGAN~\cite{zhu2017unpaired} to alleviate the domain gap between gallery faces and hallucinated ones.

\subsubsection{Evaluation}
The performance comparisons of VividGAN and other face hallucination methods on both under-controlled and unconstrained databases are shown in Tab.~\ref{tablefr}.
We take the rank-1 and rank-5 recognition rates as the evaluation metrics.
As indicated by Tab.~\ref{tablefr}, the face recognition rates of our hallucinated frontal HR faces are superior to those of the non-frontal LR faces and TANN's results on all databases.
This demonstrates that our VividGAN achieves remarkable identity preservation ability and the promising potential on challenging real-world identification problems, such as drone-based face recognition, video surveillance face recognition, etc.}}

\subsection{Comparisons with SOTA on Face Expression Classification}
{\color{black}{In this section, we manifest that our VividGAN also benefits low-resolution face expression classification tasks.}}

\begin{figure*}[ht]
\centering
\includegraphics[height=4.2cm]{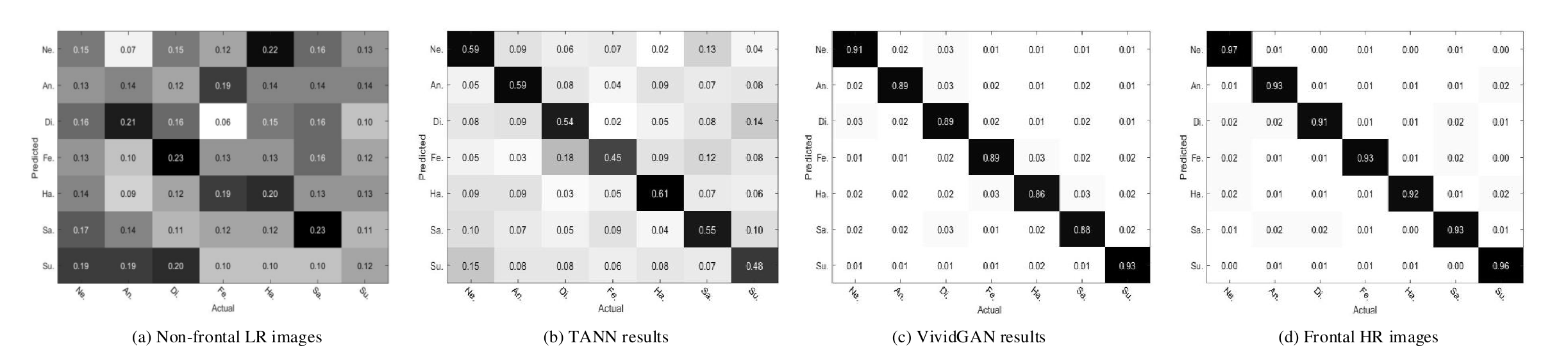}
\caption{The confusion matrices between the true labels and different predicted labels on the MMI database.}
\label{figfer}
\vspace{-0.2cm}
\end{figure*}

\subsubsection{Experimental Settings}
We perform a standard 10-fold subject-independent cross-validation~\cite{liu2014facial,liu2014feature} on the MMI database.
First, the synthesized frontal/non-frontal MMI face pairs are split into 10 subsets according to the identity information and the individuals in any two subsets are mutually exclusive.
In each experiment, 9 subsets are used for training and the remaining one subset for testing.
We train all the compared hallucination models on the same training dataset and employ a state-of-the-art expression classification model, VGG-VD-16~\cite{simonyan2014very} to identify the facial expression of HR faces hallucinated from non-frontal LR ones.
Here, state-of-the-art face hallucination methods are used to upsample the testing faces, while the classification results of the non-frontal LR faces upsampled by bicubic interpolation and the ground-truth frontal HR faces are also provided as baselines.
At last, the expression classification performance for each method are obtained by averaging the results of the 10 folds, as indicated in Tab.~\ref{tablefr}.

\subsubsection{Evaluation}
{\color{black}{As implied in Tab.~\ref{tablefr}, the hallucinated face images of our VividGAN are more authentic to the ground-truth frontal HR faces in comparison to the state-of-the-art, thus leading to superior face expression classification rates.
Particularly, the face expression classification rate of our hallucinated faces exceeds that of the input LR ones by a large margin of 65.07$\%$.
Fig.~\ref{figfer} illustrates the corresponding confusion matrices.
These matrices indicate that VividGAN recovers photo-realistic facial details more faithfully, and is more practical for the low resolution face expression classification task.}}

\begin{table}[htb]
\caption{Facial expression classification results for different methods on the MMI database}
\centering
\begin{tabular}{@{}c|c|cc@{}}
\toprule
\multirow{2}{*}{SR method} & \multicolumn{2}{c}{Accuracy} \\ \cmidrule(l){2-3} 
                           & F+SR          & SR+F          \\ \midrule
Bicubic                    & 26.08\%       & 27.95\%       \\ 
SRGAN~\cite{ledig2017photo}                      & 31.26\%       & 33.02\%       \\ 
CBN~\cite{zhu2016deep}                        & 29.30\%       & 30.17\%       \\ 
WaveletSRnet~\cite{huang2019wavelet}               & 35.58\%       & 37.34\%       \\ 
TDAE~\cite{yu2017face}                       & 34.72\%       & 35.69\%       \\ \midrule
TANN~\cite{yu2019can}             & \multicolumn{2}{c}{56.70\%}\\ \midrule
Non-frontal LR             & \multicolumn{2}{c}{23.46\%}  \\ \midrule
Frontal HR                 & \multicolumn{2}{c}{94.62\%}  \\ \midrule
VividGAN                   & \multicolumn{2}{c}{\textbf{88.53\%}}  \\ \bottomrule
\end{tabular}
\label{tablefsr}
\vspace{-0.3cm}
\end{table}

\subsection{User Study}
\begin{figure}[htb]
\centering
\includegraphics[height=4.2cm]{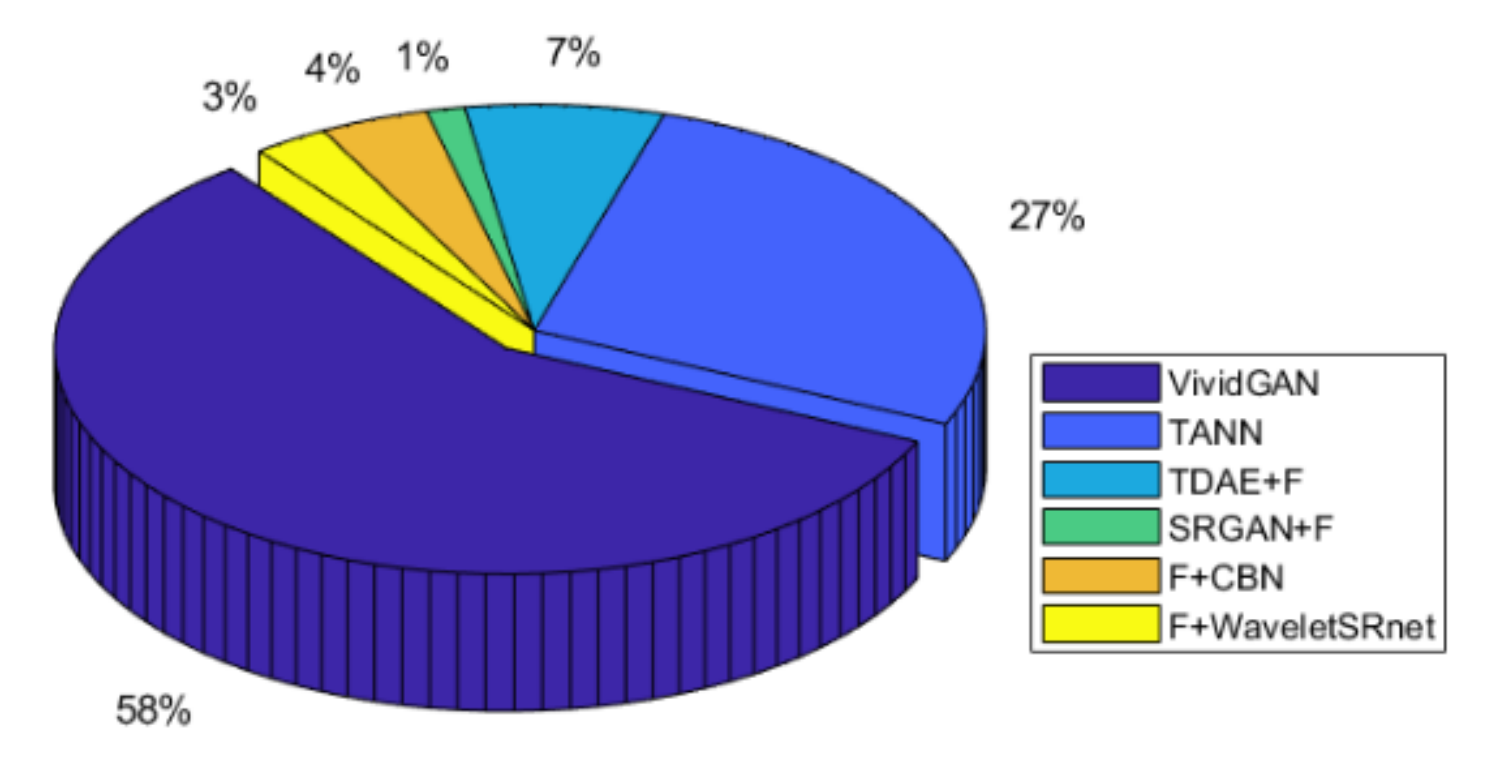}
\caption{User study. Results of our VividGAN and other state-of-the-art methods on CelebA faces.}
\label{figuse}
\vspace{-0.4cm}
\end{figure}

{\color{black}{We conduct a user study to further demonstrate the effectiveness and superiority of VividGAN.}}
Specifically, we choose fifty female and fifty male LR faces under various poses from CelebA as our testing samples.
For each testing sample, six different hallucinated HR frontal face images are shown at the same time to a user, of which one is generated by VividGAN and others are generated by the competing methods, \emph{i.e.,} TANN~\cite{yu2019can} and other four combination methods.
Each participant is required to choose the most similar one with respect to the ground-truth image.
We invite twenty participants to accomplish our user study.
After the voting, we collect 2000 votes from the participants and show the percentage of votes for each compared method in Fig.~\ref{figuse}.
The result shows that the hallucinated faces obtained by our VividGAN are more favored than other methods.

\section{Conclusion}
{\color{black}{This paper lodges an innovative ``preview" and ``revise" mechanism to hallucinate a face blueprint and touch up this blueprint
for enriching its HR facial details.
Specifically, we present an end-to-end trainable VividGAN framework to streamline the two processes, hallucination and frontalization on tiny non-frontal face images, in a coarse-to-fine fashion.
Moreover, to compensate for the obscure details in vital facial components, we also provide a solution to recovering vivid and fine-grained facial components directly from non-frontal LR faces.
VividGAN fully utilizes the facial prior knowledge and achieves face SR (an upscaling factor of $8\times$) along with frontalization (with pose variations up to $90^{o}$).
Experimental results validate the effectiveness of VividGAN, which yields identity-preserving faces and substantially boosts the performance of downstream tasks, \emph{i.e.,} face recognition and expression classification.}}

\section*{Acknowledgment}
This work was supported by the National Natural Science Foundation of China (No.61871123) and Key Research and Development Program in Jiangsu Province (No.BE2016739).
This work was supported in part by ARC under Grant DP180100106 and DP200101328.

\ifCLASSOPTIONcaptionsoff
  \newpage
\fi

\bibliographystyle{IEEEtran}
\bibliography{IEEEabrv,VividGAN}

\end{document}